\documentclass[lettersize,journal]{IEEEtran}

\usepackage{amsmath,amsfonts}
\usepackage{algorithmic}
\usepackage{algorithm}
\usepackage{amsmath,amssymb,amsthm}

\newtheorem{theorem}{Theorem}

\newtheorem{proposition}{Proposition}

\theoremstyle{definition}

\theoremstyle{remark}

\usepackage{array}
\usepackage[caption=false,font=normalsize,labelfont=sf,textfont=sf]{subfig}
\usepackage{textcomp}
\usepackage{stfloats}
\usepackage{url}
\usepackage{verbatim}
\usepackage{graphicx}
\usepackage{cite}
\usepackage{mathtools}
\usepackage{cite}
\usepackage{amsmath,amssymb,amsfonts}
\usepackage{algorithmic}
\usepackage{graphicx}
\usepackage{textcomp}
\usepackage{xcolor}
\def\BibTeX{{\rm B\kern-.05em{\sc i\kern-.025em b}\kern-.08em
    T\kern-.1667em\lower.7ex\hbox{E}\kern-.125emX}}
\usepackage{siunitx}
\usepackage{url}

\usepackage[hidelinks]{hyperref}
\begin{document}

\title{MARCH: Morphable Affine Robot with Cellular Hierarchy}

\author{Eric Lewis, Mile Chen, and Hossein Rastgoftar,~\IEEEmembership{Member,~IEEE,}
 \thanks{The authors are with the Aerospace Engineering Department, the University of Arizona, Tucson, USA 85719 Email: \{etlewis,~milechen, hrastgoftar\}@arizona.edu.}
}

\markboth{
}%
{Shell \MakeLowercase{\textit{et al.}}: A Sample Article Using IEEEtran.cls for IEEE Journals}


\maketitle

\begin{abstract}
This paper presents an affine-transformation framework for coordinating the deformation of a morphable ground vehicle composed of mechanically interconnected powered and unpowered cells. Three boundary cells specify the desired global affine motion, while interior-cell reference positions are defined through fixed barycentric combinations of neighboring cell positions over a hierarchical interaction graph. By leveraging the affine invariance of barycentric coordinates, the prescribed global deformation is mapped directly into boundary-cell position references and inter-cell linkage angle references. A seven-cell prototype integrates three omnidirectional mobile robots, four unpowered interior cells, and nine actuated interconnections. The implementation combines optical motion tracking with feedback control of boundary-cell motion and interconnection angles. Experiments evaluate prescribed scaling and shear sequences by comparing measured arm angles and cell trajectories with their references. The observed behavior highlights the effects of joint friction, wheel slip, and actuator limitations on deformation tracking. The resulting platform provides a physical testbed for investigating embodied intelligence through the interaction of reconfigurable morphology, mechanical coupling, sensing, and feedback control.
\end{abstract}

\vspace{-0.3cm}
\section{Introduction}
\label{sec:introduction}
Affine transformations (ATs) offer a scalable approach to collective motion, allowing multi-agent systems to translate, rotate, scale, and shear through a low-dimensional description of their configuration~\cite{mazmanyan2026decentralized,rastgoftar2016continuum,rastgoftar2021scalable}. In $\mathbb{R}^{n}$, $n+1$ affinely independent leaders specify the desired transformation, while followers determine their reference positions through local interactions with low computational cost. Suitable bounds on the singular values of the transformation Jacobian preserve separation between finite-sized agents, while obstacle-aware planning allows the team to reshape for narrow passages and constrained environments. These capabilities motivate extending AT-based coordination to morphable ground vehicles, where mechanically interconnected powered and unpowered cells must realize a prescribed global deformation through local actuation and feedback.

Building on these principles, we introduce the Morphable Affine Robot with Cellular Hierarchy (MARCH), a proof-of-concept ground vehicle comprising $N$ mechanically interconnected cells arranged in a deformable planar structure. MARCH gives the leader--follower architecture of AT-based multi-agent coordination a physical realization through three correspondences: powered boundary cells serve as leader agents and prescribe the global deformation; unpowered interior cells serve as follower agents whose reference positions are determined by local geometric relationships; and actuated intercell linkages provide the physical counterparts of virtual inter-agent communication links. Whereas communication links convey the information used to coordinate separate agents, MARCH's mechanical linkages transmit forces and constrain relative motion to realize coordinated deformation. A hierarchical interaction graph specifies the reference dependencies among cells, while sensing and feedback regulate boundary motion and linkage actuation, connecting local coordination rules to the controlled deformation of a single mechanically coupled robot.
    \subsection{Related Work}

        Coordinating large-scale robotic collectives through low-dimensional continuum abstractions has been widely studied in multi-agent formation control. By prescribing global ATs via a small subset of leader agents, follower agents can track combinations of translation, rotation, scaling, and shear using local barycentric coordinates or stress-matrix consensus~\cite{rastgoftar2016continuum, zhao2018affine, rastgoftar2021scalable, mazmanyan2026decentralized}. However, existing affine formation frameworks operate almost exclusively over virtual communication graphs connecting physically independent vehicles. Embodying these continuum deformation laws within a single mechanically coupled ground structure requires translating virtual graph edges into physical linkage constraints that transmit forces and position unpowered payload modules.

        On the mechanical side, shape-changing terrestrial platforms have largely been explored through modular self-reconfigurable robots (MSRRs) such as SMORES-EP~\cite{liu2019distributed}, polygon-based origami platforms such as Mori~\cite{belke2017mori}, reconfigurable wheel-limbed rovers such as NASA's ATHLETE~\cite{wilcox2007athlete}, and statistical ``particle robotics'' clusters of loosely coupled, radially expanding disc modules~\cite{li2019particle}. Whereas stochastic particle clusters lack deterministic internal pose control and traditional MSRRs require full actuation and sensing at every module, MARCH introduces a hybrid active--passive cellular hierarchy in which three powered boundary vehicles coordinate an interior lattice of unpowered payload-carrying cells through deterministic closed-chain triangulation.

        Parallel efforts in morphable aerial and multibody vehicles highlight the kinematic trade-offs of articulated shape reconfiguration~\cite{morphing_quadrotors_literature_review, morphology_adaptation_flight_performance}. Planar morphing platforms employ Sarrus, scissor, or retractable ring linkages~\cite{sarrus_linkage_single_dof_drone, x_morf_drone, ring_rotor_drone, eagle_claw_drone} for aperture traversal and whole-body grasping, whereas nonplanar and multibody architectures~\cite{overactuated_quadrotor, hofa_transformable_uav, transformable_linear_planar_multirotor, dragon_nonplanar_transforming_drone} interconnect multiple propelled links via servo-actuated joints. Because serial multibody chains suffer from controllability degradation and structural deflection near singular geometries~\cite{transformable_linear_planar_multirotor}, extending morphable architectures to closed-chain planar truss topologies offers a compelling pathway to increase structural rigidity and distribute payload support across passive interior nodes.
\vspace{-0.3cm}

\subsection{Contributions}

Unlike conventional affine formation control of
physically independent agents and morphable robots
based primarily on serial or open-chain mechanisms
~\cite{transformable_linear_planar_multirotor,
dragon_nonplanar_transforming_drone},
MARCH physically realizes hierarchical affine
coordination within a closed-chain, mechanically
coupled active--passive robotic structure.
Global affine transformations are translated into
coordinated boundary-cell motion and linkage
actuation, enabling controlled deformation of
passive interior cells. The main contributions are:

\begin{itemize}
    \item \textbf{Hierarchical affine coordination:}
    A geometry-based framework mapping global
    affine transformations into boundary-cell
    trajectories and linkage-angle references,
    enabling coordinated deformation of passive
    interior cells through mechanical coupling.

    \item \textbf{Closed-chain mechatronic architecture:}
    Design and implementation of a seven-cell
    morphable robot comprising three powered
    omnidirectional boundary cells, four passive
    payload-carrying interior cells, and nine
    actuated linkages for coordinated scaling
    and shear.

    \item \textbf{Experimental characterization:}
    Quantitative assessment of affine deformation
    through measured cell-position and linkage-angle
    errors, identifying the effects of kinematic
    overconstraint, joint friction, and actuator
    saturation on reconfiguration performance.
\end{itemize}

MARCH adopts a kinematic framework rather than
a full multibody dynamic model, which would require
explicit treatment of inertial coupling, contact
forces, friction, and actuator dynamics. The objective
is to generate geometrically feasible affine
transformations and achieve final-configuration
acquisition, without requiring exact transient
tracking. Low-friction mechanisms and smooth,
low-speed trajectories facilitate reconfiguration
through position- and angle-feedback control.
Although unmodeled dynamics introduce transient
errors, experimental results demonstrate practical
final-configuration acquisition within the tested
operating conditions, without claiming general
dynamic feasibility or stability guarantees.

\vspace{-0.3cm}
\subsection{Outline}

The remainder of this paper is organized as follows.
Section~\ref{sec:problem_statement} defines the
affine deformation control problem.
Section~\ref{sec:concept} introduces MARCH's
hierarchical architecture, affine transformation
framework, and reference configuration.
Section~\ref{sec:at_synthesis} develops formal
reconfiguration specifications, geometric feasibility
conditions, and safety-constrained AT synthesis.
Section~\ref{sec:experimental_setup} presents
the mechanical design, control implementation,
and hardware architecture.
Section~\ref{sec:results} evaluates experimental
scaling and shear performance.
Section~\ref{sec:discussion} examines mechanical
constraints and performance limitations.
Finally, Section~\ref{sec:conclusion} summarizes
the findings and future research directions.

\vspace{-0.3cm}
\section{Problem Statement}
\label{sec:problem_statement}

Consider a planar morphable ground vehicle comprising
$N$ circular structural modules (\emph{cells}) of
uniform radius $R$, indexed by
$\mathcal{V}=\{1,\ldots,N\}$.
The cells are partitioned into three actively driven
boundary cells, $\mathcal{V}_B=\{1,2,3\}$, defining
a nondegenerate triangular domain, and $N_I=N-3$
unpowered, payload-carrying interior cells,
$\mathcal{V}_I=\{4,\ldots,N\}$, located within its
convex hull. Unlike conventional multi-agent formations coordinated
through wireless communication links
~\cite{mazmanyan2026decentralized,rastgoftar2021scalable},
MARCH mechanically couples its cells through a
closed-chain network of single-degree-of-freedom
(DOF) rotary linkages connected to freely rotating
perimeter slider rings. Coordinating this
active--passive mechanical architecture presents
three coupled engineering challenges:

        \begin{enumerate}
            \item \emph{Hierarchical Continuum Coordination:} Prescribing a six-DOF global AT $\boldsymbol{\eta}(t)\in\mathbb{R}^6$ (planar translation, rigid-body rotation, axial scalings, and symmetric shear) and mapping it consistently into desired boundary-vehicle trajectories $\mathbf{p}_i(t)$ ($i\in\mathcal{V}_B$) and inter-cell linkage joint angles $\theta_{ij}^d(t)$ ($(i,j)\in\mathcal{E}_a$ for every two connected cells $i$ and $j$, where cell interconnections are formally defined by feedforward interaction graph $\mathcal{G}_s(\mathcal{V},\mathcal{E}_s)$, with edge set $\mathcal{E}_s\subset \mathcal{V}\times \mathcal{V}$.
            \item \emph{Workspace-Constrained Reference Design:} Selecting a nondegenerate reference configuration $\{\mathbf{a}_i\}_{i\in\mathcal{V}}$ and admissible parameter bounds such that bidirectional shear and anisotropic scaling can be executed without violating the finite reach bounds of the physical linkages.
            \item \emph{Minimum-State Closed-Loop Execution:} Regulating global lattice deformation using minimum online state feedback---measuring only the three boundary-vehicle poses and nine inter-cell joint angles, without direct position feedback from the unpowered interior cells---so that kinematic triangulation physically drives the passive interior cells to their desired affine trajectories.
        \end{enumerate}

\begin{figure}[t]
    \centering
    \includegraphics[width=0.49\textwidth]{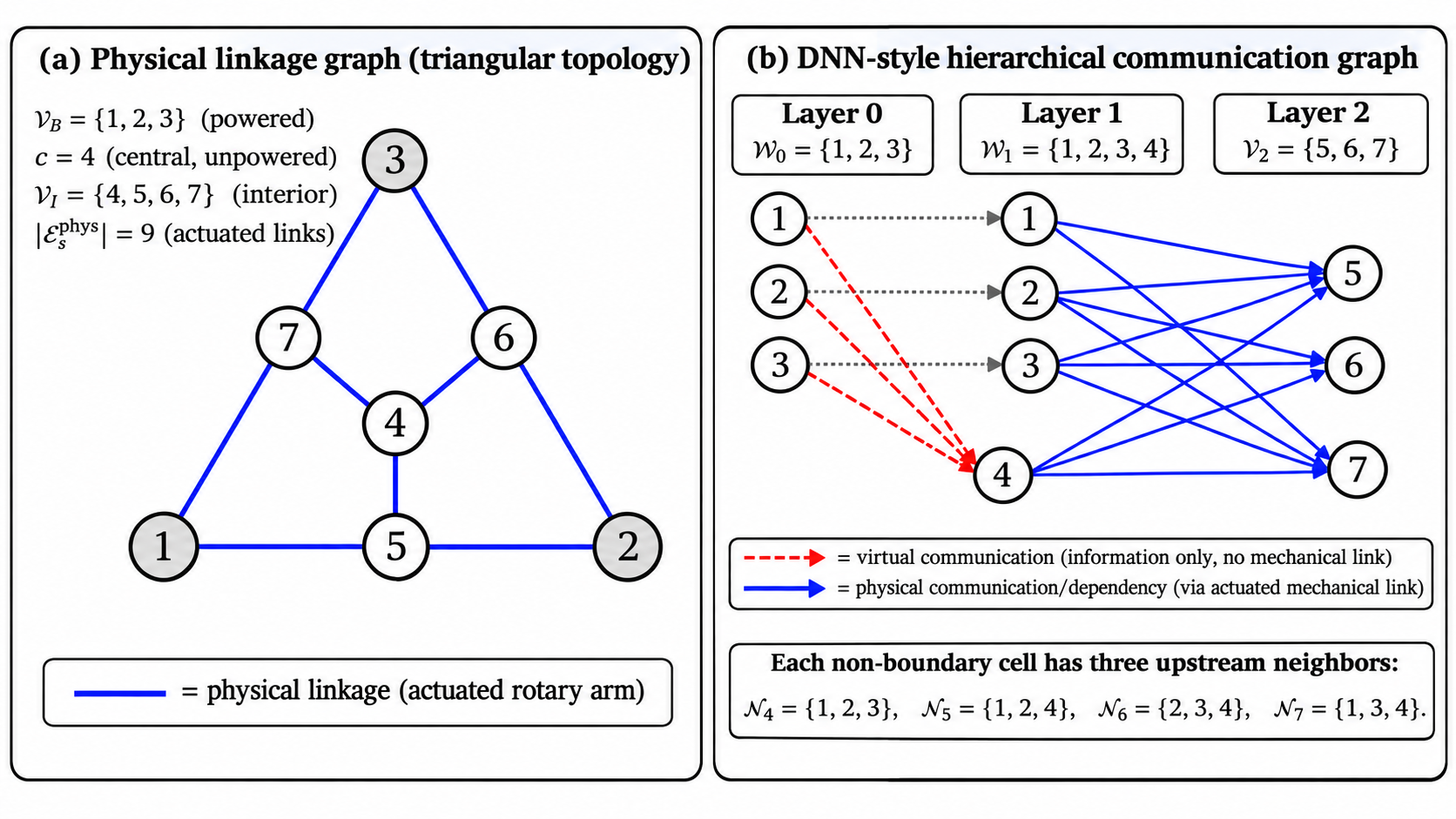}
    \vspace{-0.5cm}
    \caption{Seven-cell MARCH architecture: (a) triangular physical linkage topology comprising three powered boundary cells, four unpowered interior cells, and nine actuated linkages; (b) DNN-inspired hierarchical coordination topology with cumulative upstream cell sets. Red dashed arrows represent virtual communication between the powered boundary cells and central cell \(c=4\), while solid blue arrows represent physical linkage dependencies governing cells \(5\), \(6\), and \(7\), each with three upstream neighbors.}
    \label{fig:march_hierarchy}
\end{figure}

The objective of this work is to design, implement, and experimentally characterize a seven-cell MARCH prototype ($N_B=3$ and $N_I=4$) that realizes closed-chain affine morphing and quantifies its physical
tracking accuracy. MARCH employs a hierarchical architecture that integrates global affine coordination, virtual information exchange, and physical mechanical interconnections. As illustrated in Fig.~\ref{fig:march_hierarchy}, the hierarchical architecture connects virtual information dependencies, physical mechanical interconnections, and experimental feedback within a unified framework for affine morphing.

\noindent\textbf{Assumption 1:}
The cell centers undergo planar affine deformation,
while individual mechanical linkages are permitted
to articulate in different planes. It is assumed
that sufficient out-of-plane articulation is
available to resolve potential linkage-to-linkage
and linkage-to-cell interference during
reconfiguration.
\vspace{-0.3cm}
\section{Scientific Foundation}
\label{sec:concept}
       In our graph-theoretic formulation, each physical cell $i\in\mathcal{V}$ corresponds to a \emph{node} (or \emph{vertex}) of the interaction graph (analogous to an individual \emph{agent} in multi-agent formation control~\cite{mazmanyan2026decentralized,rastgoftar2021scalable}), while the mechanical linkages connecting adjacent cells correspond to graph edges.

\vspace{-0.3cm}

\subsection{Hierarchical Cell Architecture and Communication}
\label{subsec:hierarchical_topology}

The cell set $\mathcal V$ is partitioned into $M+1$ disjoint
layers indexed by $\mathcal M:=\{0,\ldots,M\}$:
\begin{equation}
\label{eq:cell_partition}
    \mathcal V=\bigcup_{l=0}^{M}\mathcal V_l,
    \qquad
    \mathcal V_l\cap\mathcal V_q=\emptyset,
    \quad l\neq q,
\end{equation}
where $\mathcal V_0=\mathcal V_B$ contains the powered boundary
cells, $\mathcal V_1=\{c\}$ contains the unpowered central cell,
and $\mathcal V_2,\ldots,\mathcal V_M$ contain the remaining
unpowered, payload-carrying interior cells. Accordingly,
$\mathcal V_I=\bigcup_{l=1}^{M}\mathcal V_l$ denotes the set
of interior cells. Inspired by the feedforward architecture of deep neural
networks \cite{10824818}, the cumulative cell set through layer $l$ is
defined as
\begin{equation}
\label{eq:W_l}
    \mathcal W_l
    =
    \bigcup_{q=0}^{l}\mathcal V_q,
    \qquad l\in\mathcal M.
\end{equation}
The hierarchical information dependencies are represented
by the directed graph
$\mathcal G_s=(\mathcal V,\mathcal E_s)$, where
\begin{equation}
\label{eq:MGV_edges}
\begin{aligned}
    \mathcal E_s
    &=\mathcal E_s^{\mathrm{virt}}
      \cup\mathcal E_s^{\mathrm{phys}},\\
    \mathcal E_s^{\mathrm{virt}}
    &\subseteq \mathcal V_0\times\mathcal V_1,\\
    \mathcal E_s^{\mathrm{phys}}
    &\subseteq
    \bigcup_{l=2}^{M}
    \left(\mathcal W_{l-1}\times\mathcal V_l\right),
\end{aligned}
\end{equation}
with
$\mathcal E_s^{\mathrm{virt}}\cap
\mathcal E_s^{\mathrm{phys}}=\emptyset$.
Virtual edges represent information exchange between
the powered boundary cells and the central cell without
mechanical connections. In contrast, physical edges
represent information dependencies associated with
actuated mechanical linkages between upstream and
downstream cells. For each non-boundary cell $i$, the upstream neighbor
set is defined by
\begin{equation}
\label{eq:upstream_neighbors}
    \mathcal N_i
    =
    \left\{
    j\in\mathcal V:(j,i)\in\mathcal E_s
    \right\},
    \qquad
    |\mathcal N_i|=3,
    \quad i\in\mathcal V\setminus\mathcal V_0.
\end{equation}
Thus, each non-boundary cell depends on exactly three
upstream neighbors, establishing a feedforward hierarchy
that propagates the prescribed deformation from the
powered boundary cells to the interior cells.

For the seven-cell MARCH prototype ($N=7$, $M=2$),
the hierarchy consists of three layers:
\begin{equation}
\label{eq:seven_cell_layers}
    \mathcal V_0=\{1,2,3\},
    \qquad
    \mathcal V_1=\{4\},
    \qquad
    \mathcal V_2=\{5,6,7\}.
\end{equation}
The three powered boundary cells communicate virtually
with the unpowered central cell $c=4$, while the
remaining interior cells are interconnected with their
upstream neighbors through nine actuated rotary arms.
The corresponding virtual and physical edge sets are
\begin{subequations}
\label{eq:seven_cell_edges}
\begin{align}
    \mathcal E_s^{\mathrm{virt}}
    &=
    \{(1,4),(2,4),(3,4)\},
    \label{eq:seven_cell_virtual_edges}\\
    \mathcal E_s^{\mathrm{phys}}
    &=
    \big\{
    (1,5),(1,7),(2,5),(2,6),
    \nonumber\\
    &\qquad
    (3,6),(3,7),(4,5),(4,6),(4,7)
    \big\}.
    \label{eq:seven_cell_physical_edges}
\end{align}
\end{subequations}
Consequently, the upstream neighbor sets are $\mathcal N_4=\{1,2,3\}$, $\mathcal N_5=\{1,2,4\}$, $\mathcal N_6=\{2,3,4\}$, and $\mathcal N_7=\{1,3,4\}$.

This architecture preserves the three-neighbor
dependency of the feedforward hierarchy while
distinguishing virtual information exchange from
physical mechanical interconnections.
\vspace{-0.3cm}
\subsection{MARCH Affine Transformation}
The desired AT of the MARCH is given by
\begin{equation}
\label{ATOriginal}
    \mathbf{p}_i
    =
    \mathbf{Q}\mathbf{a}_i+\mathbf{d},
    \qquad i\in\mathcal{V},
\end{equation}
where $\mathbf{a}_i\in\mathbb{R}^2$ is the fixed \emph{reference position} of cell $i$, $\mathbf{p}_i\in\mathbb{R}^2$ is its \emph{global desired position}, $\mathbf{Q}\in\mathbb{R}^{2\times2}$ defines the deformation of the reference configuration, and $\mathbf{d}=\left[d_1~d_2\right]^\top\in\mathbb{R}^2$ specifies its translation. To provide independent, direct control over rigid-body rotation, symmetric planar shear, and axial scaling, we decompose $\mathbf{Q}$ as
\begin{equation}
\label{DECOM}
    \mathbf{Q}
    =
    \mathbf{R}(\theta)\mathbf{U},
\end{equation}
\begin{equation}
\label{eq:U_matrix}
    \mathbf{U}
    =
    \mathbf{S}(\gamma)
    \operatorname{diag}(\lambda_1,\lambda_2)
    =
    \begin{bmatrix}
    \lambda_1 & \lambda_2\tan{\frac{\gamma}{2}} \\
    \lambda_1\tan{\frac{\gamma}{2}} & \lambda_2
    \end{bmatrix},
\end{equation}
where $\mathbf{R}(\theta)\in\mathrm{SO}(2)$ represents rigid-body rotation by angle $\theta$, $\mathbf{S}(\gamma)$ is the symmetric shear matrix with unit diagonal entries and off-diagonal entries $\tan(\gamma/2)$, $\lambda_1,\lambda_2>0$ are the axial scaling factors along the reference $x$- and $y$-axes, and $\gamma\in(-\pi/2,\pi/2)$ is the symmetric shear angle skewing both axes by $\gamma/2$ toward $y=x$. Characterizing the AT by six decision variables gives the compact form
\begin{equation}
\label{eq:global_desired_position}
    \boldsymbol{\eta}=\begin{bmatrix}
        d_1&d_2&\lambda_1&\lambda_2&\theta&\gamma
    \end{bmatrix}^\top,
    \qquad
    \mathbf{p}_i
    =
    \mathbf{f}\!\left(\boldsymbol{\eta},\mathbf{a}_i\right),
    \quad i\in\mathcal{V}.
\end{equation}
\begin{figure}[htbp]
\centerline{\includegraphics[width=0.98\linewidth, keepaspectratio]{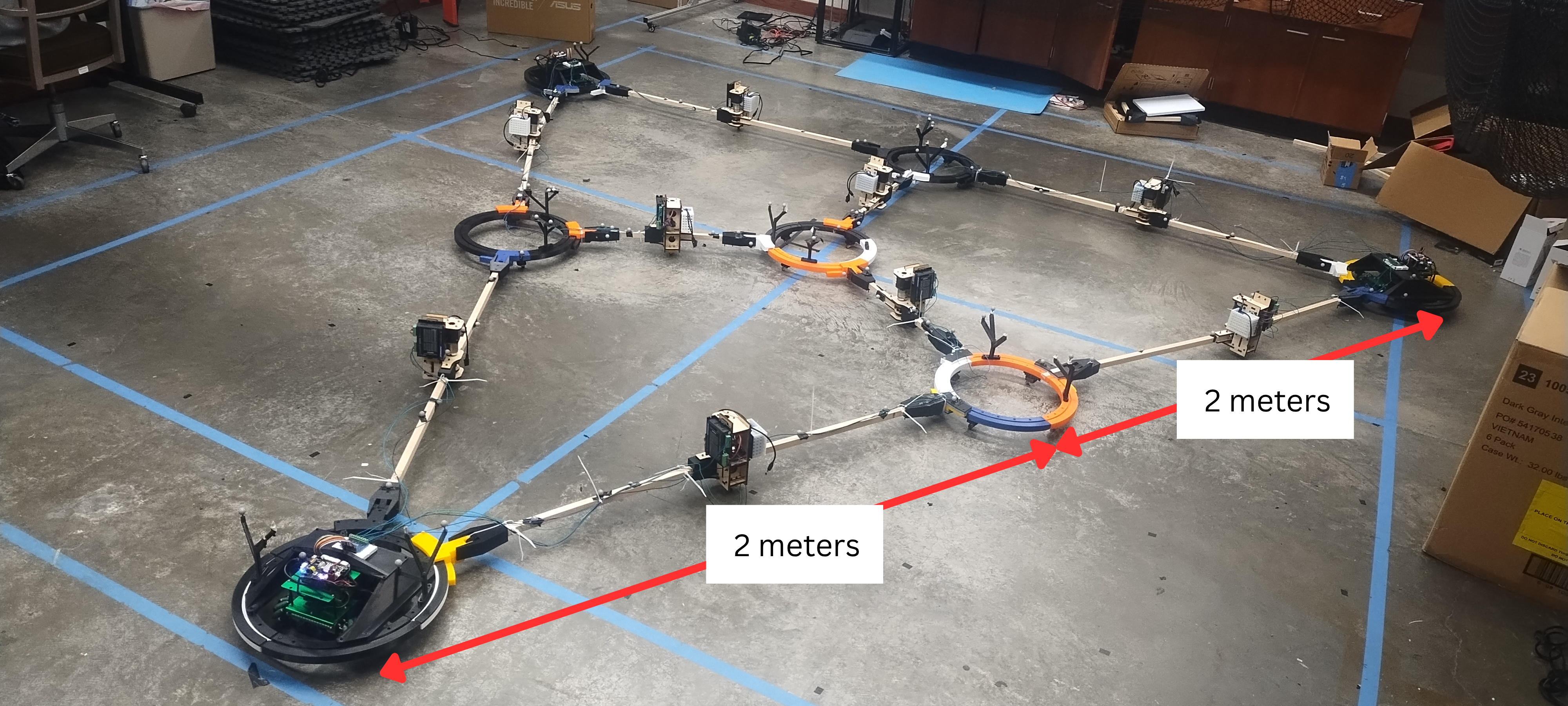}}
\caption{MARCH fully-extended configuration with side length $D=4m$.
}
\label{fig:max_scaling_affine}
\end{figure}

\vspace{-0.3cm}

\subsection{Reference and Fully Extended Configurations}
\label{subsec:reference_configuration}

Let $\mathbf a_i\in\mathbb R^2$ denote the reference
position of cell $i\in\mathcal V$. The three boundary
cells occupy the vertices of an equilateral triangle,
while the interior-cell positions satisfy the
hierarchical barycentric relations
\begin{equation}
\label{referencecomweights}
\mathbf a_i=
\sum_{j\in\mathcal N_i}w_{i,j}\mathbf a_j,
\quad
\sum_{j\in\mathcal N_i}w_{i,j}=1,
\quad w_{i,j}>0,
\end{equation}
for $i\in\mathcal V\setminus\mathcal V_0$.
The reference configuration is feasible and
nondegenerate but need not correspond to maximum
physical extension.

The fully extended configuration, illustrated in
Fig.~\ref{fig:max_scaling_affine}, forms an equilateral
boundary triangle of side length $D$. Let
$\mathbf b_i\in\mathbb R^2$ denote the corresponding
cell positions. For the seven-cell MARCH topology,
cell $4$ occupies the centroid, while cells $5$, $6$,
and $7$ occupy the boundary-edge midpoints:
\begin{equation}
\label{eq:extended_barycentric}
\begin{aligned}
\mathbf b_4&=\frac{\mathbf b_1+\mathbf b_2+\mathbf b_3}{3},\\
\mathbf b_5&=\frac{\mathbf b_1+\mathbf b_2}{2},\\
\mathbf b_6&=\frac{\mathbf b_2+\mathbf b_3}{2},\\
\mathbf b_7&=\frac{\mathbf b_1+\mathbf b_3}{2}.
\end{aligned}
\end{equation}

\begin{figure}[htbp]
\centerline{\includegraphics[width=0.96\linewidth, keepaspectratio]{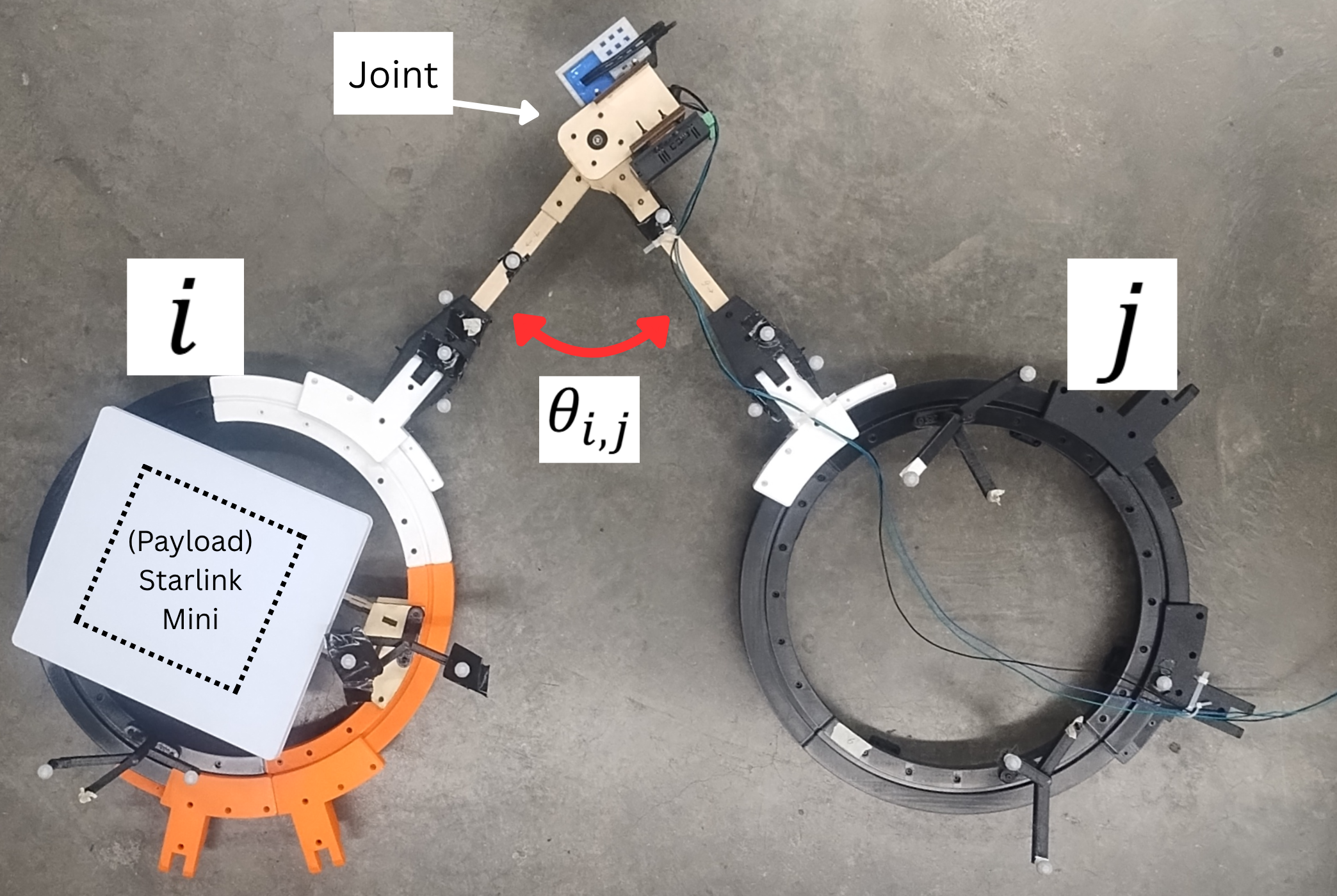}}
\caption{Physical layout of cell connection mechanism.}
\label{fig:physical_cell_connection_mechanism}
\end{figure}

Each physical linkage
$(i,j)\in\mathcal E_s^{\mathrm{phys}}$
consists of two equal-length bars of length $L_{i,j}$
connected by an actuated elbow joint
(Fig.~\ref{fig:physical_cell_connection_mechanism}).
For uniform cell radius $R$, full extension
($\theta_{i,j}=\pi$) yields
\begin{equation}
\label{eq:reference_bar_length}
L_{i,j}=
\frac{\|\mathbf b_i-\mathbf b_j\|-2R}{2}.
\end{equation}

\begin{proposition}
\label{prop:march_bar_lengths}
For the fully extended seven-cell MARCH topology
defined by \eqref{eq:extended_barycentric}, the six
outer and three inner linkages have respective bar
lengths
\begin{equation}
\label{eq:bar_lengths}
L_{\mathrm{outer}}=\frac{D}{4}-R,
\qquad
L_{\mathrm{inner}}=\frac{D}{4\sqrt{3}}-R.
\end{equation}
\end{proposition}

\begin{proof}
By symmetry and the centroid--midpoint geometry
of an equilateral triangle, the outer and inner
linkage spans are
\begin{equation}
\begin{aligned}
\ell_{\mathrm{outer}}
&=\|\mathbf b_5-\mathbf b_1\|=\frac{D}{2},\\
\ell_{\mathrm{inner}}
&=\|\mathbf b_5-\mathbf b_4\|
=\frac{D}{2\sqrt{3}}.
\end{aligned}
\end{equation}
Substitution into \eqref{eq:reference_bar_length}
yields \eqref{eq:bar_lengths}.
\end{proof}

The physical linkage dimensions are therefore
independent of the selected reference configuration,
with positive bar lengths requiring $D>4\sqrt{3}R$.

\section{Safety-Constrained Affine Transformation Synthesis}
\label{sec:at_synthesis}

The MARCH configuration is parameterized by
$\boldsymbol{\eta}=
[d_1,d_2,\lambda_1,\lambda_2,\theta,\gamma]^\top$.
The reference configuration is
$\boldsymbol{\eta}_{\mathrm{ref}}
=[0,0,1,1,0,0]^\top$,
corresponding to $\mathbf Q=\mathbf I_2$ and
$\mathbf d=\mathbf 0$.
Let $\boldsymbol{\eta}_0$ and
$\boldsymbol{\eta}_f$ denote the initial and
prescribed final configurations, respectively.
The desired configuration $\boldsymbol{\eta}(t)$
is generated by the affine transformation (AT)
in \eqref{ATOriginal}, with
$\boldsymbol{\eta}(T_0)=\boldsymbol{\eta}_0$.

The actual configuration $\boldsymbol{\eta}_a(t)$
is reconstructed from the measured boundary-cell
positions $\mathbf r_i(t)=[x_i(t),y_i(t)]^\top$,
$i=1,2,3$, satisfying
\begin{equation}
\mathbf r_i(t)=\mathbf{f}\left(\boldsymbol{\eta}_a(t),\mathbf{a}_i\right)
=\hat{\mathbf Q}(t)\mathbf a_i
+\hat{\mathbf d}(t).
\end{equation}
For three noncollinear reference boundary cells,
define
\begin{equation}
\mathbf A_0=
\begin{bmatrix}
\mathbf a_1^\top&1\\
\mathbf a_2^\top&1\\
\mathbf a_3^\top&1
\end{bmatrix}.
\end{equation}
The actual AT parameters are obtained from
\begin{equation}
\begin{bmatrix}
\hat Q_{11}&\hat Q_{21}\\
\hat Q_{12}&\hat Q_{22}\\
\hat d_1&\hat d_2
\end{bmatrix}
=
\mathbf A_0^{-1}
\begin{bmatrix}
x_1&y_1\\
x_2&y_2\\
x_3&y_3
\end{bmatrix}.
\end{equation}
Decomposition of $\hat{\mathbf Q}$ according
to \eqref{DECOM}--\eqref{eq:U_matrix},
on a fixed admissible branch, yields
\begin{equation}
\boldsymbol{\eta}_a(t)=
[\hat d_1,\hat d_2,\hat\lambda_1,
\hat\lambda_2,\hat\theta,\hat\gamma]^\top.
\end{equation}
\subsection{Formal Reconfiguration Specifications}
\label{subsec:safety}

Linear Temporal Logic (LTL) specifies continuous
safety and eventual final-configuration acquisition,
where $\Box$ and $\Diamond$ denote the \emph{always}
and \emph{eventually} operators, respectively.

\noindent\textbf{1) Liveness:}
For a prescribed tolerance $\delta>0$, define
\begin{equation}
\label{eq:march_goal_predicate}
\pi_{\mathrm{goal}}(t):
\bigwedge_{i\in\mathcal V}
\left(
\|\mathbf r_i(t)-
\mathbf f(\boldsymbol{\eta}_f,\mathbf a_i)\|
\leq\delta
\right).
\end{equation}
Eventual acquisition and retention of the final
configuration require
\begin{equation}
\label{eq:march_ltl_liveness}
\varphi_{\mathrm{live}}
=\Diamond\Box\,\pi_{\mathrm{goal}}.
\end{equation}

\noindent\textbf{2) Collision Avoidance:}
Cell collision avoidance requires
\begin{equation}
\label{eq:march_ltl_collision}
\varphi_{\mathrm{col}}
=
\Box\left(
\bigwedge_{\substack{i,j\in\mathcal V\\i<j}}
\|\mathbf r_i(t)-\mathbf r_j(t)\|
\geq2R
\right).
\end{equation}

\noindent\textbf{3) Mechanical Feasibility:}
Let $\theta_{ij}(t)$ denote the actual elbow angle,
$\theta_{\min}=\pi/3$ its minimum admissible value,
and $\pi_{\mathrm{link}}(t)$ the collision-free
linkage predicate, excluding prescribed mechanical
connections. Mechanical feasibility requires
\begin{equation}
\label{eq:march_mechanical_ltl}
\varphi_{\mathrm{mech}}
=
\Box\left[
\bigwedge_{(i,j)\in\mathcal E_s^{\mathrm{phys}}}
\left(
\theta_{\min}\leq\theta_{ij}(t)\leq\pi
\right)
\land\pi_{\mathrm{link}}(t)
\right].
\end{equation}
The overall reconfiguration specification is
\begin{equation}
\label{eq:march_ltl_objective}
\varphi=
\underbrace{
\varphi_{\mathrm{col}}\land
\varphi_{\mathrm{mech}}
}_{\varphi_{\mathrm{safe}}}
\land\varphi_{\mathrm{live}}.
\end{equation}

\subsection{Geometric Feasibility and Admissible Configurations}
\label{subsec:geometric_feasibility}

Define the minimum reference cell spacing as
$d_{\min}=\min_{i\neq j}
\|\mathbf a_i-\mathbf a_j\|>2R$.
Since
\begin{equation}
\|\mathbf p_i-\mathbf p_j\|
\geq
\sigma_{\min}(\mathbf Q)
\|\mathbf a_i-\mathbf a_j\|,
\end{equation}
where
$\sigma_{\min}(\mathbf Q)
=\sqrt{\lambda_{\min}(\mathbf Q^\top\mathbf Q)}$,
bounded position-tracking errors
$\|\mathbf r_i-\mathbf p_i\|\leq\varepsilon$
for all $i\in\mathcal V$ imply that a sufficient
condition for robust cell collision avoidance is
\cite{mazmanyan2026decentralized,rastgoftar2021scalable}
\begin{equation}
\label{eq:march_robust_safety}
\sigma_{\min}(\mathbf Q)
\geq\frac{2(R+\varepsilon)}{d_{\min}}.
\end{equation}
Setting $\varepsilon=0$ recovers the nominal
cell-separation condition.

\begin{theorem}
\label{prop:scaling_bounds}
Let the reference boundary cells form an equilateral
triangle of side length $D_{\mathrm{ref}}$, with one
edge parallel to the $y$-axis. Let $D$ denote the
fully extended boundary-triangle side length and
prescribed maximum boundary-edge length. Define
\begin{equation}
\rho=\frac{D}{D_{\mathrm{ref}}},
\qquad
\tau=\tan\frac{\gamma}{2}.
\end{equation}
For $\lambda_1,\lambda_2>0$ and $|\gamma|<\pi/2$,
boundary-edge feasibility holds if and only if
\begin{equation}
\label{eq:scaling_bounds}
\begin{aligned}
0<\sigma_{\min}(\mathbf Q)
\leq\lambda_1
&\leq\lambda_{1,\max}
=\frac{2\rho}{\sqrt{3}}\cos\frac{\gamma}{2},\\
0<\sigma_{\min}(\mathbf Q)
\leq\lambda_2
&\leq\lambda_{2,\max}
=\rho\cos\frac{\gamma}{2},
\end{aligned}
\end{equation}
and the coupled scaling--shear constraint
\begin{equation}
\label{eq:coupled_scaling_bound}
\frac{1+\tau^2}{4}
(3\lambda_1^2+\lambda_2^2)
+\sqrt{3}|\tau|\lambda_1\lambda_2
\leq\rho^2
\end{equation}
is satisfied.
\end{theorem}

\begin{proof}
The reference boundary-edge directions are
\begin{equation}
\mathbf v_1=
\begin{bmatrix}0\\1\end{bmatrix},
\quad
\mathbf v_{2,3}=
\begin{bmatrix}\sqrt{3}/2\\\pm1/2\end{bmatrix}.
\end{equation}
Since $\mathbf Q=\mathbf R(\theta)\mathbf U$
and $\mathbf R^\top\mathbf R=\mathbf I$,
boundary-edge feasibility is equivalent to
$\|\mathbf U\mathbf v_k\|^2\leq\rho^2$,
$k=1,2,3$. Direct substitution yields
\begin{equation}
\begin{aligned}
\|\mathbf U\mathbf v_1\|^2
&=(1+\tau^2)\lambda_2^2,\\
\|\mathbf U\mathbf v_{2,3}\|^2
&=\frac{1+\tau^2}{4}
(3\lambda_1^2+\lambda_2^2)
\pm\sqrt{3}\tau\lambda_1\lambda_2.
\end{aligned}
\end{equation}
Enforcing these inequalities gives
\eqref{eq:coupled_scaling_bound} and
\begin{equation}
\frac{3}{4}(1+\tau^2)\lambda_1^2\leq\rho^2,
\quad
(1+\tau^2)\lambda_2^2\leq\rho^2,
\end{equation}
establishing the upper bounds in
\eqref{eq:scaling_bounds}.

For the lower bounds, choosing
$\mathbf z=[1,-\tau\lambda_1/\lambda_2]^\top$
gives
\begin{equation}
\sigma_{\min}(\mathbf Q)
\leq\frac{\|\mathbf U\mathbf z\|}{\|\mathbf z\|}
=
\frac{\lambda_1(1-\tau^2)}
{\sqrt{1+\tau^2\lambda_1^2/\lambda_2^2}}
\leq\lambda_1.
\end{equation}
By symmetry, $\sigma_{\min}(\mathbf Q)\leq\lambda_2$.
Conversely, \eqref{eq:scaling_bounds} and
\eqref{eq:coupled_scaling_bound} imply
$\|\mathbf U\mathbf v_k\|^2\leq\rho^2$
for $k=1,2,3$, completing the proof.
\end{proof}

\noindent\textbf{1) Linkage Feasibility:}
For each $(i,j)\in\mathcal E_s^{\mathrm{phys}}$,
define the desired and maximum linkage spans as
\begin{equation}
\label{eq:linkage_distances}
\begin{aligned}
\ell_{ij}(\boldsymbol{\eta})
&=\|\mathbf Q(\boldsymbol{\eta})
(\mathbf a_j-\mathbf a_i)\|,\\
\ell_{ij}^{\max}
&=\|\mathbf b_j-\mathbf b_i\|
=2(L_{ij}+R).
\end{aligned}
\end{equation}
Under the radial-alignment assumption, the desired
elbow angle for feasible linkage spans is
\begin{equation}
\label{thijd}
\theta_{ij}^d(\boldsymbol{\eta})
=
2\sin^{-1}\left(
\frac{\ell_{ij}(\boldsymbol{\eta})}
{\ell_{ij}^{\max}}
\right).
\end{equation}
The joint limits are enforced by
\begin{equation}
\label{eq:march_link_safety}
\pi_{\mathrm{arm}}^d(\boldsymbol{\eta})
=
\bigwedge_{(i,j)\in\mathcal E_s^{\mathrm{phys}}}
\left(
\sin\frac{\theta_{\min}}{2}
\leq
\frac{\ell_{ij}(\boldsymbol{\eta})}
{\ell_{ij}^{\max}}
\leq1
\right).
\end{equation}
\noindent\textbf{2) Admissible Configuration Sets:}
Using Theorem~\ref{prop:scaling_bounds}, define
the scaling--shear feasibility predicate
\begin{equation}
\label{eq:scaling_feasibility}
\begin{aligned}
\pi_{\mathrm{scale}}(\boldsymbol{\eta})
={}&
\left(|\gamma|<\frac{\pi}{2}\right)
\land
\bigwedge_{i=1}^{2}
(0<\lambda_i\leq\lambda_{i,\max})\\
&\land
\left(
\frac{1+\tau^2}{4}
(3\lambda_1^2+\lambda_2^2)
+\sqrt{3}|\tau|\lambda_1\lambda_2
\leq\rho^2
\right).
\end{aligned}
\end{equation}
The admissible command set is defined as
\begin{equation}
\label{eq:march_command_set}
\mathcal C_{\mathrm{cmd}}
=
\left\{
\boldsymbol{\eta}\in\mathbb R^6:
\pi_{\mathrm{scale}}(\boldsymbol{\eta})
\right\}.
\end{equation}
To additionally enforce geometric safety, define
\begin{equation}
\label{eq:march_feasibility_predicate}
\begin{aligned}
\pi_{\mathrm{adm}}(\boldsymbol{\eta})
={}&
\left(
\sigma_{\min}(\mathbf Q(\boldsymbol{\eta}))
\geq\frac{2(R+\varepsilon)}{d_{\min}}
\right)\\
&\land\pi_{\mathrm{arm}}^d(\boldsymbol{\eta})
\land\pi_{\mathrm{link}}^d(\boldsymbol{\eta}),
\end{aligned}
\end{equation}
where $\pi_{\mathrm{link}}^d$ denotes collision-free
desired linkage geometry. The safety-constrained
configuration set is
\begin{equation}
\label{eq:march_admissible_set}
\mathcal C_\varepsilon
=
\left\{
\boldsymbol{\eta}\in\mathcal C_{\mathrm{cmd}}:
\pi_{\mathrm{adm}}(\boldsymbol{\eta})
\right\}.
\end{equation}
While $\mathcal C_{\mathrm{cmd}}$ characterizes
admissible affine commands, $\mathcal C_\varepsilon$
additionally enforces geometric safety. Continuous
desired geometric feasibility requires
\begin{equation}
\label{eq:march_desired_feasibility_ltl}
\varphi_{\mathrm{adm}}^d
=
\Box\left(
\boldsymbol{\eta}\in\mathcal C_\varepsilon
\right).
\end{equation}

\subsection{Feasible AT Trajectory Synthesis and LTL Guarantees}
\label{subsec:decentralized_AT}

Given $\boldsymbol{\eta}_0,\boldsymbol{\eta}_f
\in\mathcal C_\varepsilon$, the desired AT
configuration is generated by
\begin{equation}
\label{eq:AT_interpolation}
\boldsymbol{\eta}(t)
=(1-\beta(t))\boldsymbol{\eta}_0
+\beta(t)\boldsymbol{\eta}_f,
\end{equation}
for $t\in[T_0,T_f]$, where $T_0<T_f$ and
\begin{equation}
\label{eq:beta_function}
\beta(t)=6s^5-15s^4+10s^3,
\qquad s=\frac{t-T_0}{T_f-T_0}.
\end{equation}
The quintic interpolation satisfies
$\beta(T_0)=0$, $\beta(T_f)=1$,
$\dot\beta,\ddot\beta=0$ at both endpoints,
and $\dot\beta\geq0$.
For $t\geq T_f$,
$\boldsymbol{\eta}(t)=\boldsymbol{\eta}_f$.
The desired cell trajectories follow from
\eqref{eq:global_desired_position}.

\begin{proposition}
\label{prop:march_feasible_at}
Consider the quintic AT interpolation
in \eqref{eq:AT_interpolation}.
The desired trajectory satisfies
$\varphi_{\mathrm{adm}}^d$ if and only if
\begin{equation}
\label{eq:march_segment_feasibility}
(1-s)\boldsymbol{\eta}_0+
s\boldsymbol{\eta}_f
\in\mathcal C_\varepsilon,
\quad\forall s\in[0,1].
\end{equation}
\end{proposition}

\begin{proof}
The quintic function $\beta$ is continuous
and nondecreasing, with
$\beta(T_0)=0$ and $\beta(T_f)=1$.
Hence, $\beta([T_0,T_f])=[0,1]$.
By \eqref{eq:AT_interpolation}, the desired
trajectory traverses the entire segment
between $\boldsymbol{\eta}_0$ and
$\boldsymbol{\eta}_f$.
Therefore, $\boldsymbol{\eta}(t)\in
\mathcal C_\varepsilon$ for all
$t\geq T_0$ if and only if
\eqref{eq:march_segment_feasibility}
holds, proving $\varphi_{\mathrm{adm}}^d$.
\end{proof}

The proposition guarantees geometric feasibility
of the desired AT, but does not establish
continuous safety or final acquisition of
the physical robot. These properties are
specified by $\varphi_{\mathrm{safe}}$ and
$\varphi_{\mathrm{live}}$, respectively, and
evaluated experimentally using measured
cell positions, linkage angles, and
reconstructed actual AT parameters
$\boldsymbol{\eta}_a$.
Exact transient trajectory tracking is
not required.

\begin{figure}[htbp]
\centerline{\includegraphics[width=\linewidth, keepaspectratio]{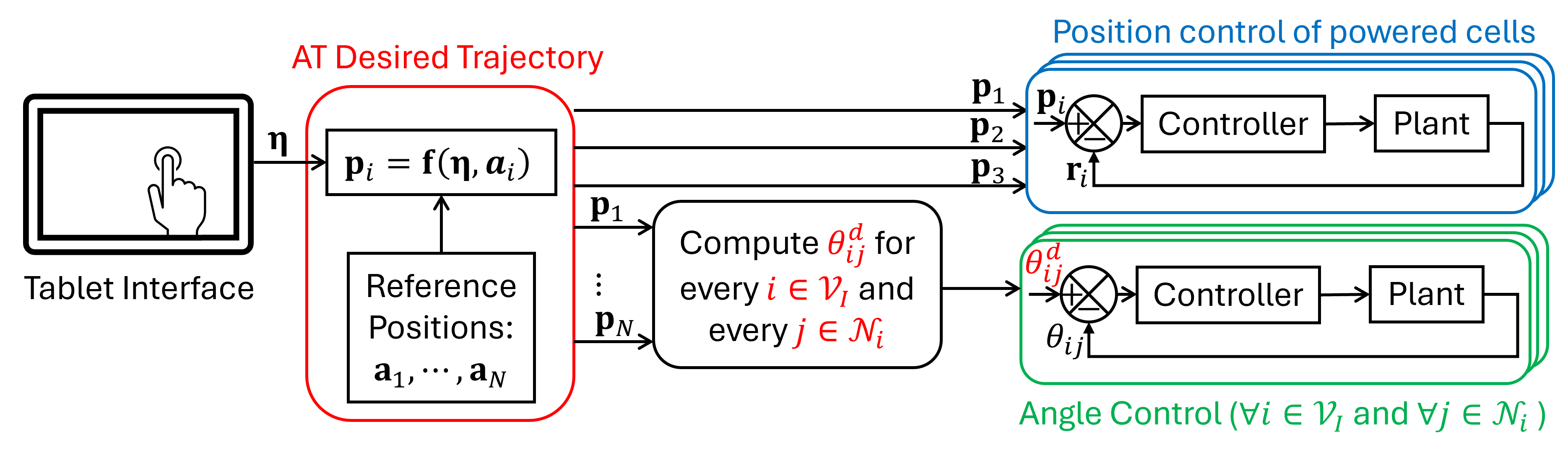}}
\caption{Proposed control architecture mapping the commanded AT parameter vector $\boldsymbol{\eta}$ into coordinated boundary-cell position references $\mathbf{p}_i$ and interior-joint angle references $\theta_{ij}^d$.}
\label{fig:ControlArchitecture}
\end{figure}
\begin{figure}[htbp]
    \centering
    \includegraphics[width=0.98\linewidth,keepaspectratio]{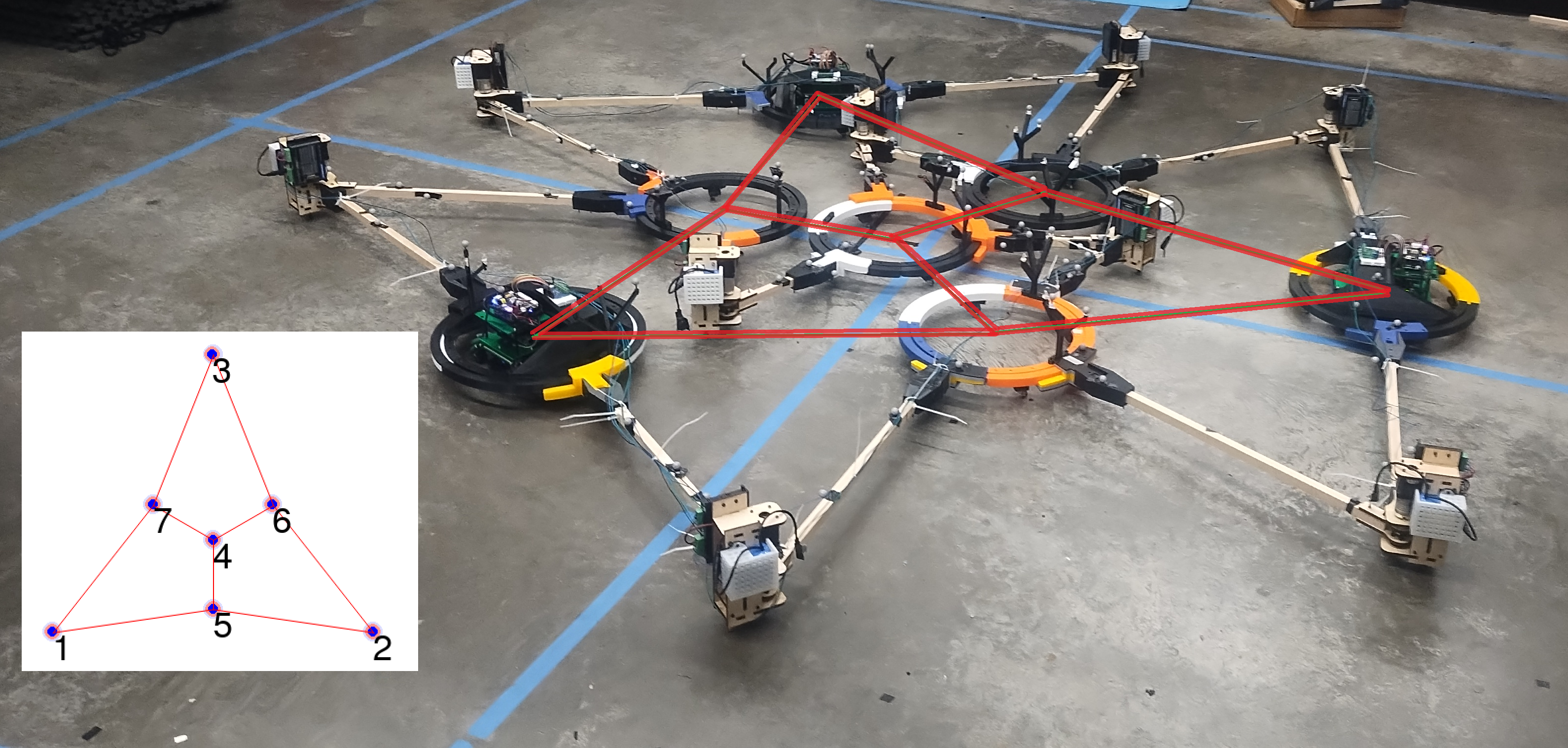}
    \caption{Intermediate implementation of the vehicle used to investigate performance on foam tile versus smooth concrete, showing the lower-section charging system. AT with $(\lambda_1,\lambda_2)=(0.50,0.50)$, illustrating aggressive scaling capability; perspective distortion affects the apparent symmetry.}
    \label{fig:tiny_affine}
\end{figure}
\vspace{-0.3cm}

\section{Engineering Realization}
\label{sec:experimental_setup}

MARCH realizes synthesized ATs through the control 
architecture shown in Fig.~\ref{fig:ControlArchitecture}, 
integrating three powered holonomic boundary cells, four 
passive interior cells, and nine stepper-actuated linkages.
Each boundary cell $i\in\mathcal V_B$ tracks its desired 
position using the error 
$\mathbf e_i(t)=\mathbf p_i(t)-\mathbf r_i(t)$, 
while each actuated linkage $(i,j)\in\mathcal E_a$ 
tracks the elbow-angle reference in 
\eqref{thijd}. A tablet-based interface specifies AT commands, and a 
motion-capture system provides feedback measurements.
Low-friction mechanical interfaces and smooth, low-speed 
deformations mitigate unmodeled dynamic effects.
Consistent with the kinematic formulation, the experiments 
emphasize final configuration convergence rather than exact 
transient tracking. To isolate scaling 
$(\lambda_1,\lambda_2)$ and shear $\gamma$, translation 
and rotation are fixed at $d_1=d_2=0$ and $\theta=0$. The following subsections describe the mechanical design, 
hardware integration, feedback control, and experimental 
evaluation.
\vspace{-0.3cm}
\subsection{Cell Interconnection Mechanism}
Each interior cell $i\in \mathcal{V}_I$ is mechanically coupled to its neighbors $j\in \mathcal{N}_i$ via the linkage assembly in Fig.~\ref{fig:physical_cell_connection_mechanism}. All seven circular cell rings have an outer radius $R=\SI{220}{\milli\meter}$ and inner radius \SI{174}{\milli\meter}, separated by a \SI{6}{\milli\meter} internal slider track. Each linkage comprises two equal-length wooden bars $L_{ij}$ joined by a central $180^\circ$ rotary elbow actuator and terminating in polymeric perimeter slider rings that constrain the bar axis normal to the cell wall while permitting unconstrained circumferential rotation. Six long outer arms (\SI{0.655}{\meter} wooden rods; effective length $L_{\text{outer}}+R=\SI{1.0}{\meter}$, \SI{2}{\meter} maximum inter-cell span) and three shorter inner arms (\SI{0.2324}{\meter} rods; $L_{\text{inner}}+R=1/\sqrt{3}\text{ m}\approx\SI{0.5774}{\meter}$, $2/\sqrt{3}\text{ m}\approx\SI{1.155}{\meter}$ maximum span) connect the lattice so that the fully extended vehicle forms an equilateral triangle of side length \SI{4}{\meter} (Fig.~\ref{fig:max_scaling_affine}), with software-enforced limits $\theta_{ij}\in[60^\circ,180^\circ]$.


The perimeter rail minimizes sliding friction by confining contact to the top planar surfaces and vertical guide walls of the 3D-printed rings. The 3D-printed end fittings joining the wooden bars to the perimeter sliders incorporate an out-of-plane revolute joint to accommodate ground irregularities and enable cell rings and elbow joints to articulate out of the plane to access underside payloads (such as the Starlink Mini terminal) and service actuators without disassembling the planar network. Both the four unpowered interior rings and nine elbow joints are supported by \SI{12.7}{\milli\meter} ($0.5\text{-in}$) low-profile swivel plate casters with 3D-printed spacer shims beneath each elbow to maintain coplanar elevation (whereas the three boundary vehicles rest on their mecanum wheels). Each elbow joint is actuated by a \SI{5}{\newton\meter} geared stepper motor (NEMA~17 with a $100:1$ planetary gearbox; exact ratio $99.55:1$; Fig.~\ref{fig:physical_cell_connection_mechanism}), whose open-loop operation provides passive overload protection by slipping magnetic poles (backdriving) under excessive internal constraint forces rather than shearing gear teeth.
\subsection{Low-Level Control Architecture}
\label{subsec:low_level_control}

MARCH AT implementation integrates onboard computation, distributed motor actuation, and external optical motion tracking. Car~1 coordinates the commanded transformation and distributes motion commands to the powered cells, while two complementary feedback mechanisms regulate boundary-cell positions and linkage angles. This subsection describes  the position- and angle-control loops used to realize coordinated deformation without directly controlling the four passive interior cells followed by the embedded hardware and communication architecture.
\subsubsection{Powered-Cell Hardware and Coordination}

Each powered boundary cell ($\mathcal{V}_B=\{1,2,3\}$) is built around a Yahboom ROSMASTER X3 omnidirectional mecanum-wheeled mobile robot equipped with an onboard Raspberry Pi~5, three StepperOnline DM542T stepper drivers, and power regulation electronics (Fig.~\ref{fig:tiny_affine}). The holonomic chassis uses its built-in \SI{12}{\volt} battery; each of the three assigned linkage-arm drivers is supplied by a dedicated \SI{12}{\volt} lithium-ion battery pack, and the onboard supplies support external charging. Directly interfacing the three stepper drivers with each Raspberry Pi~5 replaced the standalone wireless XIAO ESP32-C3 microcontrollers used in earlier prototypes. Thus, only the three boundary-cell Raspberry Pi~5 modules are required to operate the entire nine-arm lattice.

\begin{figure}[htbp]
    \centering
    \includegraphics[width=0.98\linewidth,keepaspectratio]{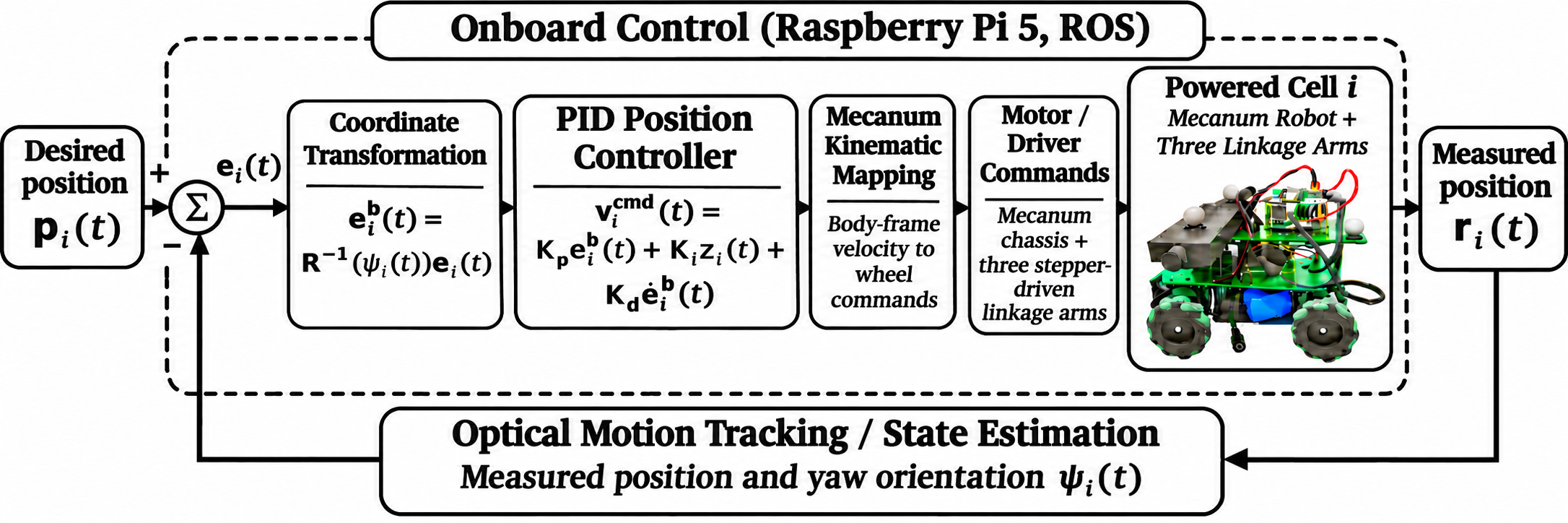}
    \caption{Closed-loop position-control architecture of powered boundary cell $i\in\mathcal V_B$. The desired position $\mathbf p_i(t)$ and Vicon-measured position $\mathbf r_i(t)$ define the tracking error $\mathbf e_i(t)$. The measured heading is used to express the error in the body-fixed mecanum frame. A PID controller with leaky integral anti-windup generates planar velocity commands, which are converted into individual wheel commands through the mecanum kinematic mapping. Optical motion tracking closes the position feedback loop.}
\label{fig:powered_cell_control}
\end{figure}

For each powered boundary cell
$i\in\mathcal V_B$, the planar tracking error is
$\mathbf e_i(t)=\mathbf p_i(t)-\mathbf r_i(t)$,
where $\mathbf p_i(t)$ and $\mathbf r_i(t)$
denote the desired and Vicon-measured positions,
respectively. Planar motion is regulated using
PID feedback with leaky integral anti-windup
and gains $(K_p,K_i,K_d)=(5.0,0.08,2.0)$.
The resulting body-frame velocity commands
are mapped to individual mecanum-wheel commands,
with optical motion tracking closing the feedback
loop (Fig.~\ref{fig:powered_cell_control}). 

Because each boundary cell's outer linkage ring
rotates freely around its perimeter, the vehicle
yaw is mechanically decoupled from the lattice
kinematics. Active yaw control is therefore
omitted to prevent rotational oscillations.
Instead, the Vicon-measured yaw angle
$\psi_i(t)$ is used to transform the
position-tracking error into the body-fixed
mecanum coordinate frame:
\begin{equation}
\mathbf{{e}}_i^b(t)
=
\mathbf{R}^{-1}(\psi_i(t))\mathbf{{e}}_i(t),
\label{eq:body_frame_error}
\end{equation}
where
\begin{equation}
\mathbf R(\psi_i)=
\begin{bmatrix}
\cos\psi_i & -\sin\psi_i\\
\sin\psi_i & \cos\psi_i
\end{bmatrix}.
\label{eq:vehicle_rotation}
\end{equation}
The commanded planar velocity is generated
using a PID position controller with leaky
integral anti-windup:
\begin{equation}
\mathbf{\mathrm{v}}_i^{\mathrm{cmd}}(t)
=
K_p\mathbf{\mathrm{e}}_i^b(t)
+
K_i\mathbf{\mathrm{z}}_i(t)
+
K_d\dot{\mathbf{\mathrm{e}}}_i^b(t)
\label{eq:boundary_pid}
\end{equation}
where $\mathbf{\mathrm{z}}_i(t)$ denotes
the integral state incorporating the implemented
leakage and anti-windup mechanism.
The proportional, integral, and derivative gains
are selected as
$(K_p,K_i,K_d)=(5.0,0.08,2.0)$.
The resulting body-frame velocity command
$\mathbf{\mathrm{v}}_i^{\mathrm{cmd}}(t)$
is converted into individual mecanum-wheel
commands through the chassis kinematic mapping.
The Vicon-measured position and orientation
are subsequently fed back to Car~1 to update
the velocity commands at each control cycle,
as illustrated in Fig.~\ref{fig:powered_cell_control}.
\begin{figure}[htbp]
    \centering
    \includegraphics[width=\linewidth,keepaspectratio]{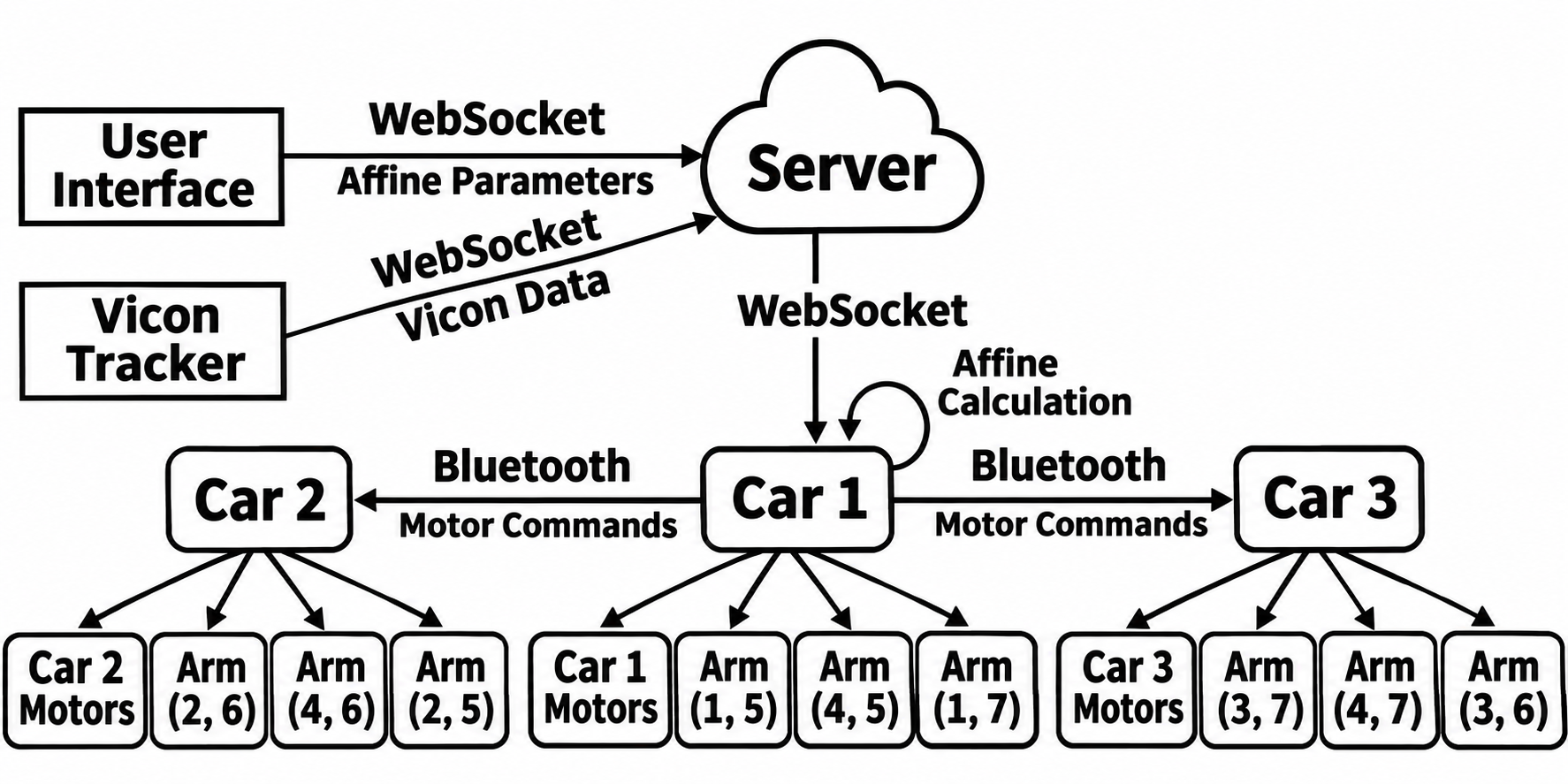}
    \caption{Software and hardware integration of the distributed MARCH architecture: user commands and motion-tracking data reach Car~1 through the WebSocket infrastructure, while Car~1 coordinates local actuation and communicates motor commands to Cars~2 and~3 via Bluetooth.}
    \label{fig:integration}
\end{figure}

\subsubsection{Integration and Feedback Control}
\label{subsubsec:feedback_control}

The software and hardware architecture is shown in
Fig.~\ref{fig:integration}. Car~1 serves as the onboard
coordinator, receiving commanded AT parameters and Vicon
measurements through an AWS EC2 WebSocket proxy.
It generates boundary-cell trajectories and nine
linkage-angle references, controls its own mecanum
chassis and stepper drivers, and transmits motor
commands to Cars~2 and~3 via Bluetooth.
Each powered cell controls its mecanum drive and
three assigned linkage actuators. 
The four passive interior cells
($\mathcal{V}_I=\{4,5,6,7\}$) are caster-supported
frames repositioned through mechanical linkages,
requiring neither onboard controllers nor independent
propulsion.

\begin{figure}[!t]
    \centering
    \includegraphics[width=0.49\textwidth,keepaspectratio]{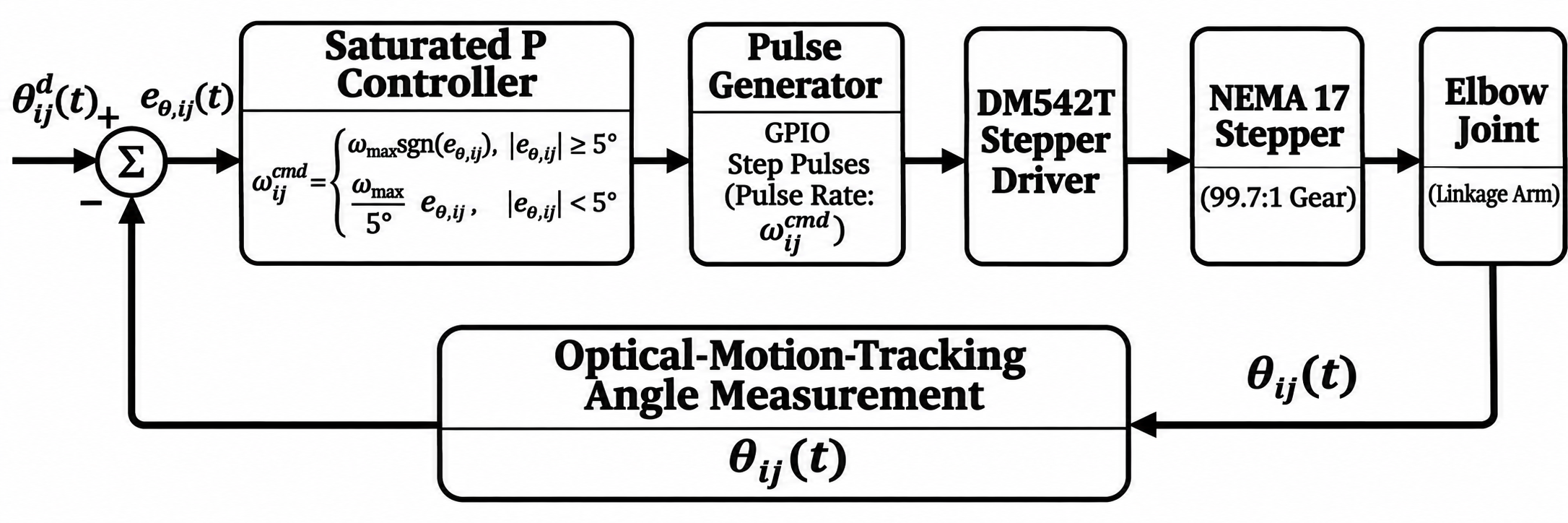}
    \caption{Closed-loop elbow-angle control architecture for each actuated linkage $(i,j)\in\mathcal{E}_s^{\rm phys}$. The reference angle $\theta_{ij}^d(t)$ is compared with the optically measured angle $\theta_{ij}(t)$. A saturated proportional controller commands a bounded angular speed, which is converted to Raspberry Pi~5 GPIO pulses, sent through a DM542T driver, and executed by a geared NEMA~17 stepper motor. Relative arm-segment yaw measurements close the feedback loop.}
    \label{fig:stepper_control}
\end{figure}

Closed-loop feedback uses the measured poses of the
three boundary cells and nine elbow angles
$\theta_{ij}(t)$, reconstructed from the relative
yaw orientations of 18 tracked arm segments.
No interior-cell position feedback is required.
The boundary-cell and elbow-angle control loops
are illustrated in Figs.~\ref{fig:powered_cell_control}
and~\ref{fig:stepper_control}, respectively. For each actuated linkage
$(i,j)\in\mathcal E_s^{\mathrm{phys}}$,
the AT planner computes the desired elbow angle
$\theta_{ij}^d(t)$ using \eqref{thijd}.
The angular tracking error is
\begin{equation}
\label{eq:angular_tracking_error}
e_{\theta,ij}(t)
=\theta_{ij}^d(t)-\theta_{ij}(t).
\end{equation}
A saturated proportional controller generates
the commanded elbow angular velocity:
\begin{equation}
\label{eq:elbow_speed_control}
\omega_{ij}^{\mathrm{cmd}}(t)=
\begin{cases}
\omega_{\max}\operatorname{sgn}(e_{\theta,ij}(t)),
& |e_{\theta,ij}(t)|\ge e_{\mathrm{th}},\\[3pt]
\displaystyle
\frac{\omega_{\max}}{e_{\mathrm{th}}}e_{\theta,ij}(t),
& |e_{\theta,ij}(t)|<e_{\mathrm{th}},
\end{cases}
\end{equation}
where $e_{\mathrm{th}}=5^\circ$ and
$\omega_{\max}\approx9.04^\circ/\mathrm{s}$.
Errors exceeding the threshold command maximum
speed, while smaller errors produce proportionally
reduced speeds.

The Raspberry Pi~5 translates the commanded speed and direction into GPIO step-pulse signals delivered to the corresponding DM542T driver. The driver energizes a geared NEMA~17 stepper motor, which rotates the elbow joint and changes the inter-cell linkage geometry. With a gearbox ratio of $99.55{:}1$,
the motor requires 19,910 half-steps
(9,955 full-step pulses) per
$180^\circ$ output rotation.
At the maximum commanded stepping rate of \SI{1}{\milli\second} per half-step, the elbow speed is approximately $\omega_{\max}=9.04^\circ/\mathrm{s}$. Expressing the command in radians per second, the corresponding half-step rate is
\begin{equation}
    f_{ij}^{\rm hs}(t)=\frac{N_{\rm hs}}{\pi}
    \left|\omega_{ij}^{\rm cmd}(t)\right|,
    \qquad N_{\rm hs}=19{,}910,
    \label{eq:stepper_pulse_rate}
\end{equation}
with the sign of $\omega_{ij}^{\rm cmd}(t)$ specifying the rotation direction and the rate bounded by the available stepping speed.

Although the stepper motors have no onboard joint encoders, optical measurement of the actual elbow angles closes the joint-level feedback loops. Car~1 updates the commands using the observed errors, including discrepancies arising from missed steps, friction, or mechanical coupling. Thus, the nine linkages are controlled using externally sensed joint angles rather than relying exclusively on open-loop pulse counting.

Together, the PID-regulated boundary cells and angle-regulated linkages drive the mechanically coupled, caster-supported interior cells toward the prescribed affine configuration. Because the closed-chain structure couples nominally separate actuators, joint friction, geometric limits, and actuator-rate saturation can produce transient errors during aggressive morphing. Figure~\ref{fig:tiny_affine} shows the converged closed-loop configuration for uniform scaling $(\lambda_1,\lambda_2)=(0.50,0.50)$ with $d_1=d_2=\theta=\gamma=0$.

 \vspace{-0.3cm}
\subsection{MARCH AT Planning through Tablet Interface}

\begin{figure}[htbp]
\centerline{\includegraphics[width=0.98\linewidth, keepaspectratio]{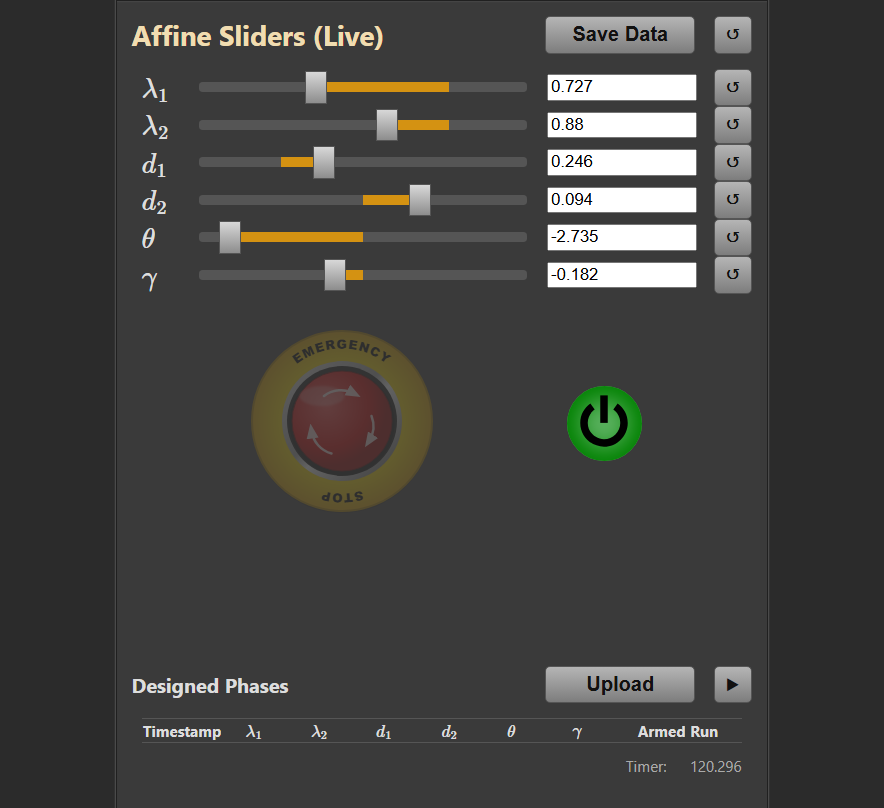}}
\caption{User interface for precise manual control and systematic testing.}
\label{fig:user_interface}
\end{figure}

The MARCH control system is accessible through a web-based tablet interface (Fig.~\ref{fig:user_interface}) that provides direct access to $\boldsymbol{\eta}=\left[d_1~d_2~\lambda_1~\lambda_2~\theta~\gamma\right]^\top$. Users can interactively command real-time deformations via parameter sliders (with reset buttons to restore reference values) or upload a preset multi-phase testing regimen interpolated by a smooth $\beta(t)$ scaling function (Section~\ref{sec:verification}). To respect the laboratory motion-tracking capture volume and linkage geometry without overextending outer arms beyond \SI{2}{\meter} or folding any elbow joint below its $60^\circ$ minimum angle limit, the interface constrains inputs to $d_1\in[-0.5,1.5]\text{ m}$, $d_2\in[-0.25,0.25]\text{ m}$, $\lambda_1,\lambda_2\in[1/2,\,2/\sqrt{3}]$, $\theta\in[-\pi,\pi]\text{ rad}$, and $\gamma\in[-1,1]\text{ rad}$. When an AT command or regimen is initiated on the tablet, target parameters $\boldsymbol{\eta}$ and timestamps are logged locally and transmitted over WebSocket via the EC2 proxy server to Car~1, which maps the global AT command into local boundary-cell position references and interior-joint angle references (Fig.~\ref{fig:ControlArchitecture}).

\vspace{-0.3cm}
    \subsection{Motion Tracking}

\begin{figure}[htbp]
        \centerline{\includegraphics[width=0.98\linewidth, keepaspectratio]{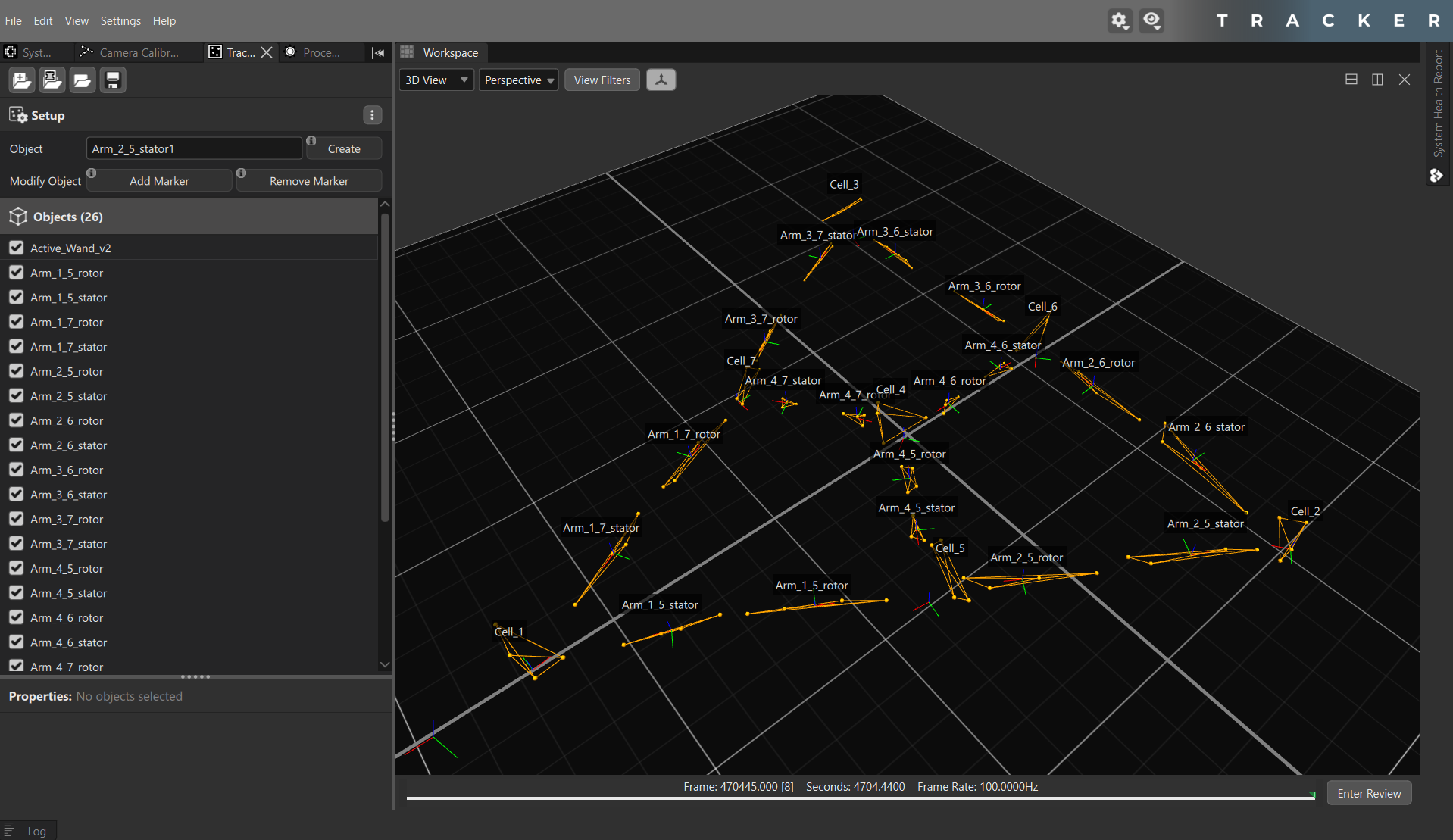}}
        \caption{All cells and arm segments are independently tracked using an optical tracking system for arm-angle measurement and cell-position validation.}
        \label{fig:vicon_implementation}
        \end{figure}

        The planar poses of all seven cells and 18 linkage arm segments are tracked using a Vicon optical motion-capture system (Vantage and Valkyrie cameras; Fig.~\ref{fig:vicon_implementation}). Each rigid body is defined by a unique four-marker constellation of \SI{14}{\milli\meter} retroreflective spheres, providing sub-centimeter positional accuracy (\SIrange{2}{3}{\milli\meter}) and yaw orientation measurements. Real-time pose streams acquired via the \verb|vicon_datastream_sdk| Python API are transmitted over JSON WebSocket to the AWS EC2 proxy, which forwards state packets to Car~1 while logging timestamped ground-truth trajectories for post-experiment verification.

        \begin{table}[htbp]
            \centering
            \caption{Phases of AT parameters.}
            \label{tab:AT_parameters}
            \setlength{\tabcolsep}{4.5pt}
            \begin{tabular}{rrrrrrr}
                \hline
                Time (s) & $d_1$ (m) & $d_2$ (m) & $\lambda_1$ & $\lambda_2$ & $\theta$ (rad) & $\gamma$ (rad) \\
                \hline
                 0.0 & 0.0 & 0.0 & 1.0 & 1.0 & 0.0 & 0.0 \\
                10.0 & 0.0 & 0.0 & 0.8 & 0.9 & 0.0 &-0.1 \\
                20.0 & 0.0 & 0.0 & 0.8 & 0.9 & 0.0 &-0.1 \\
                30.0 & 0.0 & 0.0 & 1.0 & 0.6 & 0.0 & 0.3 \\
                40.0 & 0.0 & 0.0 & 1.0 & 0.6 & 0.0 & 0.3 \\
                50.0 & 0.0 & 0.0 & 1.0 & 1.0 & 0.0 & 0.0 \\
                60.0 & 0.0 & 0.0 & 1.0 & 1.0 & 0.0 & 0.0 \\
                \hline
            \end{tabular}
        \end{table}

\vspace{-0.3cm}
\section{Experimental Proof of Concept}
\label{sec:results}
The experimental evaluation investigates affine reconfiguration within the admissible command space, including trajectories that challenge the mechanical feasibility conditions. This enables characterization of deformation tracking, actuator saturation, and mechanical recovery beyond the nominal safety-constrained operating region. The closed-loop tracking performance and structural resiliency of the seven-cell MARCH prototype were evaluated across the multi-phase regimen in Table~\ref{tab:AT_parameters}. Rather than applying external force perturbations to evaluate disturbance rejection, the prescribed trajectory drives the over-constrained lattice beyond its nominal kinematic fold limits and actuator velocity bounds. Fig.~\ref{fig:graph_stacked} illustrates the reference trajectory alongside the resulting arm angular errors and local-frame cell position errors, Fig.~\ref{fig:online_cell_paths} compares spatial cell paths against their references, and Table~\ref{tab:AT_experimental_error} summarizes the phase-by-phase root-mean-square error (RMSE) and maximum absolute error.
\vspace{-0.3cm}
\subsection{Reference Configuration and Experimental Setup}
\label{subsec:experimental_configuration}
The fully extended MARCH configuration forms an
equilateral boundary triangle of side length
$D=\SI{4}{\meter}$
(Fig.~\ref{fig:max_scaling_affine}).
To accommodate shear within linkage reach limits,
the reference boundary triangle is uniformly
contracted relative to the fully extended boundary
triangle, placing the three powered boundary cells
on a circle of radius \SI{2}{\meter}, with side length
$D_{\mathrm{ref}}=2\sqrt{3}\,\mathrm{m}
\approx\SI{3.464}{\meter}$. 
The initial boundary-cell positions coincide with
their reference positions, i.e.,
$\mathbf r_i(T_0)=\mathbf a_i$ for $i\in\mathcal V_B$. The boundary-cell reference positions are
\begin{equation}
\mathbf a_i=
2\begin{bmatrix}
\cos\left(\frac{2\pi(i-1)}{3}\right)\\
\sin\left(\frac{2\pi(i-1)}{3}\right)
\end{bmatrix},
\quad i\in\mathcal V_B.
\end{equation}
The interior-cell reference positions are determined
by the fixed hierarchical barycentric weights in
\eqref{referencecomweights}, whereas their actual
initial positions may differ, resulting in nonzero
initial tracking errors. The prescribed reconfiguration starts from and
returns to the reference affine transformation,
i.e., $\boldsymbol{\eta}_0=\boldsymbol{\eta}_f
=\boldsymbol{\eta}_{\mathrm{ref}}$. Consequently,
\begin{equation}
\mathbf Q(T_0)=\mathbf Q(T_f)=\mathbf I_2,
\qquad
\mathbf d(T_0)=\mathbf d(T_f)=\mathbf 0.
\end{equation}
A video demonstration of the experimental
reconfiguration is available
\href{https://www.youtube.com/watch?v=EaqlkJlysgk}
{online}~\cite{MARCH_video}.



\vspace{-0.3cm}
\subsection{Results}
    Following nominal baseline tracking during the initial ramp and hold (\SIrange{0}{20}{\second}), the error curves exhibit two prominent transient peaks during Phase~2 ($t\approx\SI{27}{\second}$) and Phase~3 ($t=\SI{50}{\second}$) in response to the boundary-limit maneuvers. During the Phase~2 contraction and shear ramp (\SIrange{20}{30}{\second}; $\lambda_1: 0.8\rightarrow 1.0, \lambda_2: 0.9\rightarrow 0.6, \gamma: -0.1\rightarrow 0.3$), the trajectory commands interior Arm~$(4, 7)$ beyond its minimum physical fold geometry and induces mechanical contact between Arm~$(4, 6)$ and the perimeter frame of Cell~7 ($t\approx\SI{33}{\second}$). At $t=\SI{25.2}{\second}$, the controller clamps Arm~$(4, 7)$ at its lower software limit ($\theta_{\min} = 60^\circ$), halting local contraction and producing a transient lag peaking at $t\approx\SI{27}{\second}$ ($28.37^\circ$ maximum arm error; \SI{0.393}{\meter} cell error). Throughout the \SIrange{30}{40}{\second} hold, the compliant truss accommodates this internal contact alongside stiction in Arm~$(3, 5)$ while maintaining bounded tracking errors during the hold interval (arm RMSE of $6.42^\circ$; cell RMSE of \SI{0.149}{\meter}).

    In the Phase~3 return ramp (\SIrange{40}{50}{\second}), a rapid expansion and shear reversal ($\lambda_2: 0.6\rightarrow 1.0, \gamma: 0.3\rightarrow 0.0$ over \SI{10}{\second}) exceeds the continuous speed limit of the 100:1 geared stepper motors ($\omega_{\max}\approx 9.04^\circ/\text{s}$) to evaluate closed-loop adaptation under actuator rate saturation. Specifically, the required angular rate of Arm~$(2, 5)$ exceeds $\omega_{\max}$ continuously from $t=\SI{44.0}{\second}$ to $t=\SI{47.35}{\second}$ (peaking above $11.46^\circ/\text{s}$), while Arm~$(1, 5)$ also saturates, inducing transient phase lag and open-loop magnetic pole slipping. Concurrently, the unfolding motion autonomously disengages the Arm~$(4, 6)$--Cell~7 contact ($t\approx\SI{44}{\second}$) and releases Arm~$(4, 7)$ from its $\theta_{\min}$ clamp ($t=\SI{44.35}{\second}$). Cumulative velocity lag and asymmetric boundary traction---as Car~1 repositions while Car~3 is restrained by joint and floor stiction---peak at $t=\SI{50}{\second}$ ($40.86^\circ$ maximum arm error; \SI{0.706}{\meter} cell error) before closed-loop feedback drives recovery.

    Upon entering each stationary hold window, the steep attenuation of both angular and positional error trajectories in Fig.~\ref{fig:graph_stacked}(b)--(c) demonstrates the system's ability to rapidly recover and converge following dynamic transformations, settling into narrow steady-state deadbands governed by polymeric slider-ring stiction. At initialization ($t=\SI{0}{\second}$), Arm~$(3, 6)$ exhibits a stiction offset against Cells~3 and~6, breaks free at $t\approx\SI{4}{\second}$, and re-binds at $t\approx\SI{6}{\second}$ under rising internal forces. During the final recovery window (\SIrange{50}{90}{\second}), stored strain triggers a stick-slip release at Cell~6 ($t\approx\SI{58}{\second}$), halving its position error while transiently compressing Arm~$(4, 5)$ until it clears stiction at $t\approx\SIrange{66}{68}{\second}$. By the converged final state (\SIrange{70}{90}{\second}), eight of the nine linkages converge cleanly (arm RMSE dropping from $15.82^\circ$ to $1.18^\circ$ and cell RMSE from \SI{0.231}{\meter} to \SI{0.070}{\meter}), with only Arm~$(3, 6)$ retaining a $\approx 3.6^\circ$ stiction offset---confirming that minimum-state feedback reliably restores global affine convergence after severe kinematic and rate saturation.
\vspace{-0.3cm}
    \subsection{Verification}
        \label{sec:verification}

\begin{figure}[htbp]
    \centerline{\includegraphics[width=0.86\linewidth, keepaspectratio]{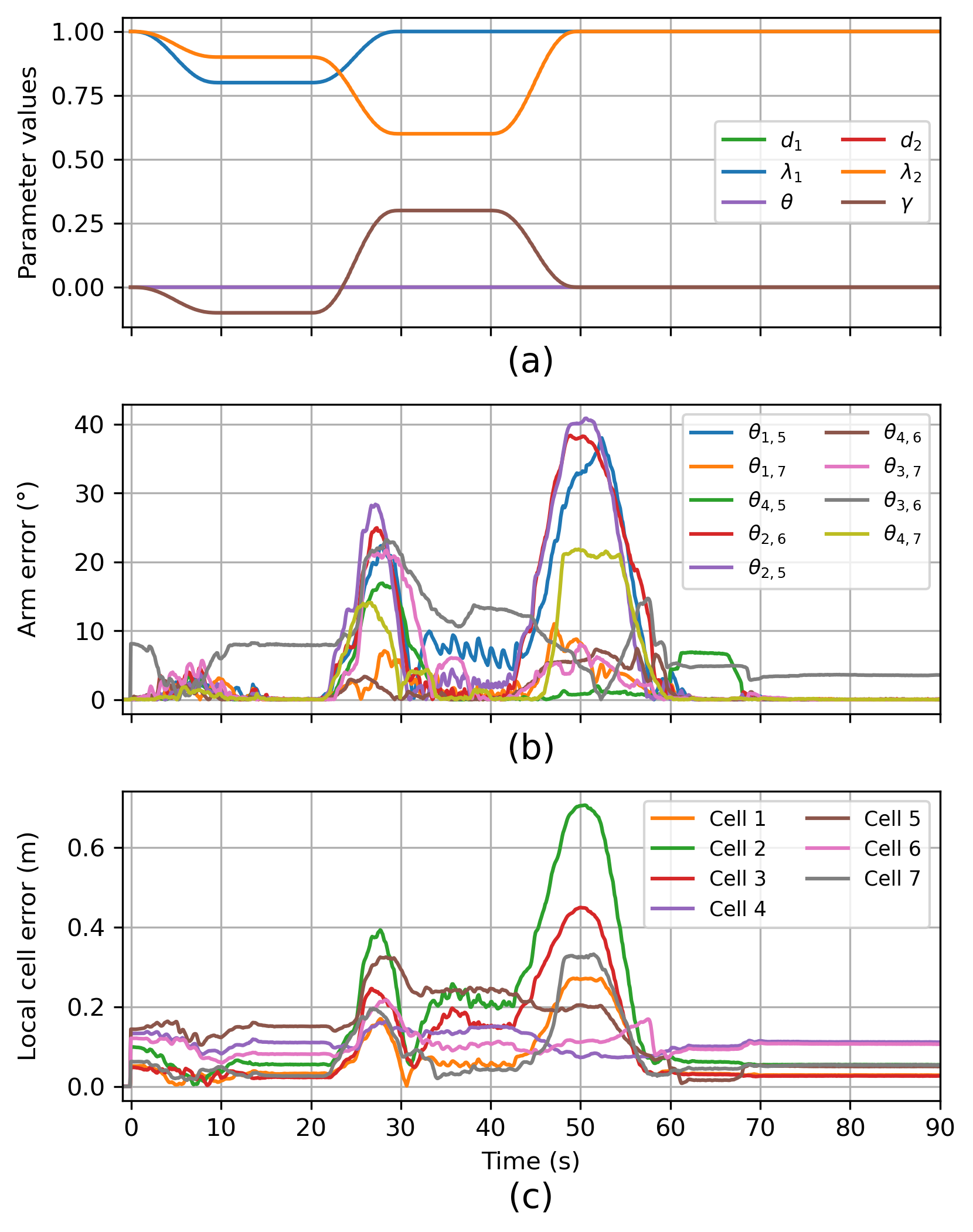}}
    \caption{Experiment results. (a) Preset AT parameter values $\boldsymbol{\eta}(t)$ generated by $\beta(r)$. (b) Absolute angular tracking error $|\theta_{ij}^d(t)-\theta_{ij}(t)|$ of the nine actuated arms. (c) Rigid-body-motion-compensated position tracking
error $\|\widetilde{\mathbf p}_i(t)-\mathbf r_i(t)\|$
of the seven cells.}
    \label{fig:graph_stacked}
    \end{figure}

\begin{figure}[htbp]
    \centerline{\includegraphics[width=0.98\linewidth, keepaspectratio]{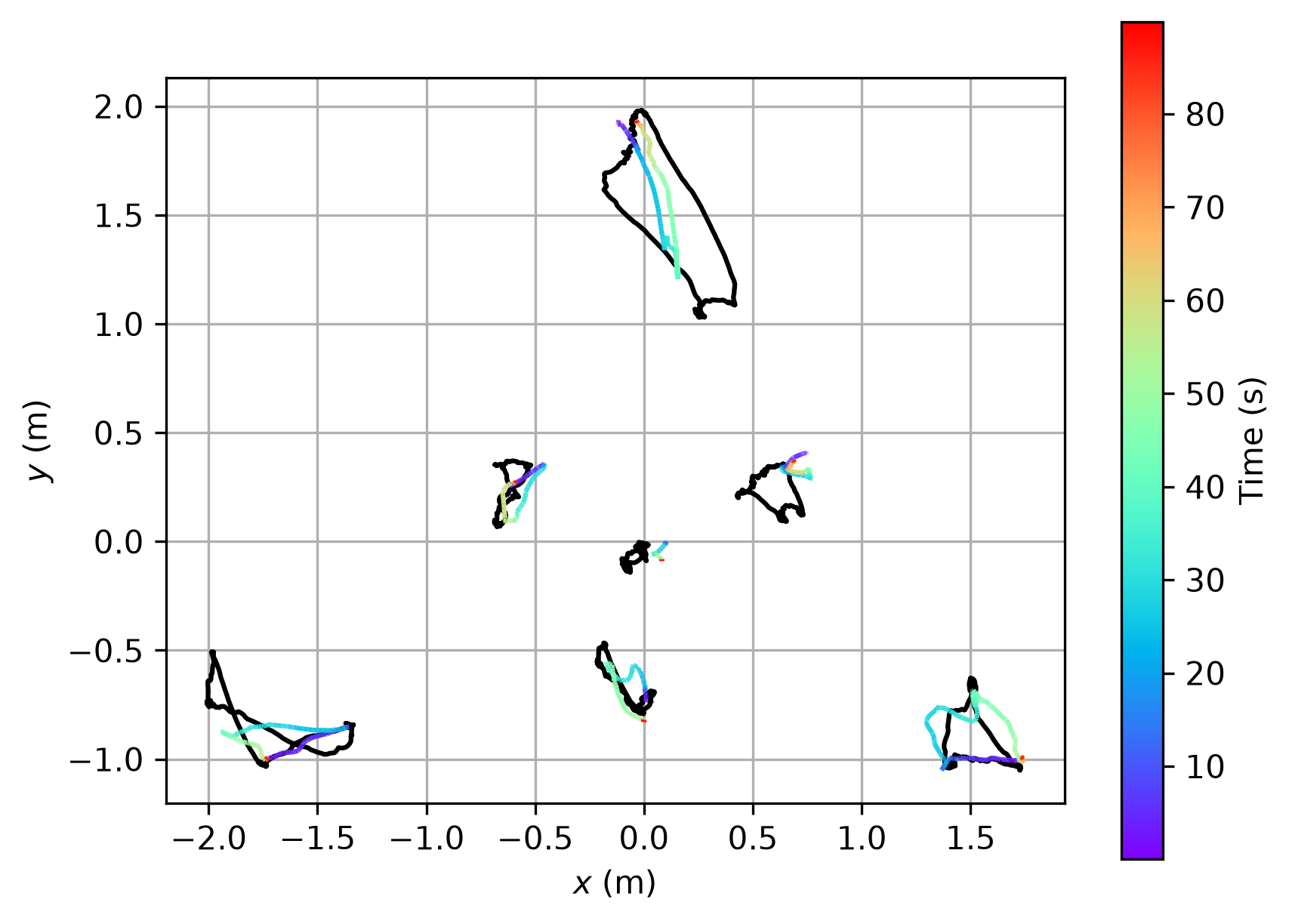}}
    \caption{Comparison of online (color gradient representing elapsed time) and desired (black) cell trajectories.}
    \label{fig:online_cell_paths}
    \end{figure}

        To evaluate deformation tracking systematically, $\boldsymbol{\eta}(t)$ is prescribed over a \SI{60}{\second} command schedule followed by a \SI{30}{\second} settling window (Table~\ref{tab:AT_parameters}), alternating \SI{10}{\second} dynamic deformation ramps with stationary hold phases. To isolate internal scaling ($\lambda_1, \lambda_2$) and shear ($\gamma$) from gross room translation, global translation and rotation are held at $d_1(t)=d_2(t)=0\text{ m}$ and $\theta(t)=0\text{ rad}$ while all three boundary cars actively track $\mathbf{p}_1(t), \mathbf{p}_2(t), \mathbf{p}_3(t)$. Consecutive waypoints in Table~\ref{tab:AT_parameters} are interpolated by using \eqref{eq:AT_interpolation}. To account for uncommanded rigid-body motion induced
by mechanical coupling and wheel slip, the realized
translation $\widehat{\mathbf d}(t)$ and rotation
$\widehat{\theta}(t)$ are reconstructed from the
measured boundary-cell positions
$\{\mathbf r_i(t)\}_{i\in\mathcal V_B}$.
The compensated desired cell positions are
\begin{equation}
\label{eq:compensated_desired_position}
\widetilde{\mathbf p}_i(t)
=
\mathbf R(\widehat{\theta}(t))
\mathbf U(t)\mathbf a_i
+\widehat{\mathbf d}(t),
\qquad i\in\mathcal V,
\end{equation}
where $\mathbf U(t)$ represents the prescribed
scaling and shear. This compensation isolates
deformation-tracking errors from rigid-body drift
in Figs.~\ref{fig:graph_stacked}(c)
and~\ref{fig:online_cell_paths}.

    \begin{table}[htbp]
        \centering
        \caption{Error metrics during Affine Transformation.}
        \label{tab:AT_experimental_error}
        \setlength{\tabcolsep}{5pt}
        \begin{tabular}{lrrrr}
            \hline
            & \multicolumn{2}{c}{Arm Error ($^\circ$)} & \multicolumn{2}{c}{Cell Error (m)} \\
            \cline{2-3} \cline{4-5}
            Phase Window (s) & RMSE & Max & RMSE & Max \\
            \hline
            0.0--10.0        &  2.34 &  8.12 & 0.086 & 0.164 \\
            10.0--20.0       &  2.67 &  8.24 & 0.081 & 0.160 \\
            20.0--30.0       & 12.21 & 28.37 & 0.167 & 0.393 \\
            30.0--40.0       &  6.42 & 21.07 & 0.149 & 0.293 \\
            40.0--50.0       & 13.63 & 40.14 & 0.243 & 0.705 \\
            50.0--60.0       & 15.82 & 40.86 & 0.231 & 0.706 \\
            60.0--70.0       &  2.46 &  6.90 & 0.064 & 0.115 \\
            70.0--90.0       &  1.18 &  3.64 & 0.070 & 0.115 \\
            \hline
            0.0--90.0 (All)  &  8.52 & 40.86 & 0.146 & 0.706 \\
            \hline
        \end{tabular}
    \end{table}

\vspace{-0.3cm}
\section{Discussion: System Analysis and Mechanical Trade-offs}
\label{sec:discussion}

        Each interior cell is constrained by three linkages ($|\mathcal{N}_i|=3$). During dynamic deformation phases, minor phase lags or velocity mismatches among the nine stepper motors generate internal constraint forces acting tangentially and radially along the cell perimeter rings. While the polymeric sliding ring joints maintain perpendicular linkage alignment, off-axis bending moments increase normal contact pressure along the FDM print surfaces, inducing nonlinear static friction (stiction). Consequently, during steady-state hold windows, residual elastic strain and joint stiction produce a bounded steady-state positional deadband rather than asymptotic zero-error convergence (Fig.~\ref{fig:graph_stacked}c).
\vspace{-0.3cm}

    \subsection{Actuator Backdrivability vs. Closed-Loop Tracking}
        A key design trade-off is the selection of open-loop \SI{5}{\newton\meter} stepper motors over high-ratio geared servomotors. Under over-constrained internal loading, rigid position-controlled servomotors risk catastrophic gear-tooth shearing; by contrast, when internal stress concentrations exceed the holding torque of the stepper motors, the rotors slip magnetic poles (skip steps), providing inherent mechanical compliance that protects the wooden linkage beams and 3D-printed interfaces. Because pole slipping invalidates open-loop step counting, closing the feedback loop via external optical tracking of the arm segments (Fig.~\ref{fig:vicon_implementation}) is essential to recover true joint-angle state estimation and maintain convergence toward $\theta_{ij}^d(t)$.
\vspace{-0.3cm}
    \subsection{Traction-to-Payload Ratio and Surface Interaction}
        For the three active boundary robots to execute gross rigid-body translation alongside internal morphing, their mecanum wheels must overcome the combined rolling and sliding resistance of the four unpowered interior cells and nine linkage assemblies against the floor. Comparing operation across laboratory floor surfaces (Fig.~\ref{fig:online_cell_paths}), high-friction foam mats increase boundary wheel traction but disproportionately increase passive structure drag, whereas smooth concrete reduces interior drag at the cost of mecanum wheel slip---motivating future low-friction payload support mechanisms.
\vspace{-0.3cm}
\section{Conclusion}
\label{sec:conclusion}

    This paper presented the design, hierarchical control formulation, and experimental characterization of MARCH, a seven-cell morphable ground vehicle that physically realizes continuum ATs through a closed-chain lattice of three active boundary robots, four unpowered payload-carrying interior cells, and nine rotary linkage arms. By exploiting the affine invariance of barycentric coordinates over a directed feedforward interaction graph, global scaling and shear deformations $\boldsymbol{\eta}(t)$ commanded through a tablet interface were mapped into boundary-vehicle trajectories and inter-cell joint-angle references using minimum state feedback. Multi-phase motion-capture experiments validated the ability of the coupled active--passive lattice to track combined anisotropic scaling and shear deformations while exposing fundamental mechatronic trade-offs among kinematic over-constraint, joint stiction, and actuator compliance. 

    The present implementation adopts a kinematic reference-generation and feedback-control framework without explicitly accounting for the coupled multibody dynamics or actuator torque constraints. While this approach successfully demonstrated the prescribed affine deformations within the experimentally tested operating envelope, the observed tracking errors during aggressive maneuvers highlight the effects of joint friction, mechanical coupling, wheel slip, and actuator-rate limitations that are not explicitly represented in the kinematic formulation. Future work will incorporate constrained multibody dynamics and actuator-aware trajectory planning to quantify internal interaction forces, evaluate torque feasibility, and improve tracking performance during faster and more demanding lattice reconfigurations.
\vspace{-0.5cm}
    \subsection{Limitations}

        Performance of the current prototype stems from opposing friction requirements: the polymeric perimeter ring sliders experience intermittent binding due to manufacturing tolerances and off-axis bending loads, whereas the mecanum boundary vehicles exhibit limited tractive authority relative to the aggregate drag of the passive structure. Additionally, while the \SI{5}{\newton\meter} stepper motors provide overload compliance by backdriving under excessive internal stress, operating them without onboard encoders requires external optical motion capture to compensate for skipped steps, and the out-of-plane revolute hinges at the bar-to-slider interfaces introduced structural compliance under combined torsional and bending loads that required mechanical reinforcement.
\vspace{-0.5cm}

    \subsection{Future Work}

        Future work will investigate alternative inter-cell actuation mechanisms that eliminate the lateral workspace footprint of planar folding elbow arms, whose outward radial projection restricts minimum inter-cell spacing and can induce link-to-cell contact during large shear maneuvers. Subsequent iterations will evaluate out-of-plane (vertical) scissor linkages, telescopic linear actuators, and low-mechanical-impedance quasi-direct-drive (QDD) proprioceptive actuators with integrated encoders~\cite{mit_cheetah_proprioceptive_design, qdd_actuator_compliant_arm} to achieve backdrivable force compliance without open-loop step loss. Furthermore, replacing large-diameter polymeric perimeter slider rings with centralized multi-coaxial pin joints equipped with onboard absolute rotary encoders will eliminate off-axis binding and enable self-contained proprioceptive state estimation without external motion capture. Reducing internal joint stiction and passive payload drag (such as for the mounted Starlink Mini terminal) will further enable a small number of powered boundary cells to manipulate larger payload networks across uneven outdoor terrains or high-order fully actuated aerial platforms~\cite{hofa_transformable_uav}.
        \vspace{-0.3cm}
\section*{Acknowledgment}
This work was supported in part by the Arizona
Board of Regents (ABOR) and Arizona Commerce
Authority (ACA) through the TRIF Regents'
Research Grant awarded to H. Rastgoftar.
\bibliographystyle{IEEEtran}
\bibliography{references}

@article{mazmanyan2026decentralized,
  title={Decentralized affine transformation for scalable and safe multi-agent aerial coordination},
  author={Mazmanyan, Garegin and Rastgoftar, Hossein},
  journal={Aerospace Science and Technology},
  volume={180},
  pages={113539},
  year={2026},
  publisher={Elsevier}
}

@book{rastgoftar2016continuum,
  title={Continuum deformation of multi-agent systems},
  author={Rastgoftar, Hossein},
  year={2016},
  publisher={Springer}
}

@ARTICLE{10824818,
  author={Rastgoftar, Hossein},
  journal={IEEE Transactions on Control of Network Systems}, 
  title={Deep and Decentralized Multiagent Coverage of a Target With Unknown Distribution}, 
  year={2025},
  volume={12},
  number={2},
  pages={1393-1405},
  doi={10.1109/TCNS.2025.3525802}}

@misc{MARCH_video,
  author       = {{SMART Laboratory}},
  title        = {{MARCH: Experimental Demonstration}},
  howpublished = {YouTube video},
  url          = {https://www.youtube.com/watch?v=EaqlkJlysgk},
  note         = {Accessed: Nov. 26, 2026}
}

@article{rastgoftar2021scalable,
  title={Scalable vehicle team continuum deformation coordination with eigen decomposition},
  author={Rastgoftar, Hossein and Atkins, Ella M and Kolmanovsky, Ilya V},
  journal={IEEE Transactions on Automatic Control},
  volume={67},
  number={5},
  pages={2514--2521},
  year={2021},
  publisher={IEEE}
}

@article{mit_cheetah_proprioceptive_design,
	title = {Proprioceptive {Actuator} {Design} in the {MIT} {Cheetah}: {Impact} {Mitigation} and {High}-{Bandwidth} {Physical} {Interaction} for {Dynamic} {Legged} {Robots}},
	volume = {33},
	copyright = {https://ieeexplore.ieee.org/Xplorehelp/downloads/license-information/IEEE.html},
	issn = {1552-3098, 1941-0468},
	shorttitle = {Proprioceptive {Actuator} {Design} in the {MIT} {Cheetah}},
	url = {http://ieeexplore.ieee.org/document/7827048/},
	doi = {10.1109/TRO.2016.2640183},
	language = {en},
	number = {3},
	urldate = {2026-08-13},
	journal = {IEEE Transactions on Robotics},
	author = {Wensing, Patrick M. and Wang, Albert and Seok, Sangok and Otten, David and Lang, Jeffrey and Kim, Sangbae},
	month = jun,
	year = {2017},
	pages = {509--522},
}

@misc{qdd_actuator_compliant_arm,
	title = {Quasi-{Direct} {Drive} for {Low}-{Cost} {Compliant} {Robotic} {Manipulation}},
	url = {http://arxiv.org/abs/1904.03815},
	doi = {10.48550/arXiv.1904.03815},
	urldate = {2026-08-13},
	publisher = {arXiv},
	author = {Gealy, David V. and McKinley, Stephen and Yi, Brent and Wu, Philipp and Downey, Phillip R. and Balke, Greg and Zhao, Allan and Guo, Menglong and Thomasson, Rachel and Sinclair, Anthony and Cuellar, Peter and McCarthy, Zoe and Abbeel, Pieter},
	month = apr,
	year = {2019},
	note = {arXiv:1904.03815 [cs.RO]},
}

@article{hofa_transformable_uav,
	title = {A novel transformable {UAV}: {HOFA} modeling, control, and experimental validations},
	volume = {173},
	issn = {0967-0661},
	shorttitle = {A novel transformable {UAV}},
	url = {https://www.sciencedirect.com/science/article/pii/S0967066126002340},
	doi = {10.1016/j.conengprac.2026.106990},
	urldate = {2026-08-13},
	journal = {Control Engineering Practice},
	author = {Lu, Shi and Tsakalis, Konstantinos and Chen, Yan},
	month = aug,
	year = {2026},
	pages = {106990},
}

@article{morphing_quadrotors_literature_review,
	title = {Morphing {Quadrotors}: {Enhancing} {Versatility} and {Adaptability} in {Drone} {Applications}—{A} {Review}},
	volume = {8},
	copyright = {http://creativecommons.org/licenses/by/3.0/},
	issn = {2504-446X},
	shorttitle = {Morphing {Quadrotors}},
	url = {https://www.mdpi.com/2504-446X/8/12/762},
	doi = {10.3390/drones8120762},
	language = {en},
	number = {12},
	urldate = {2026-08-13},
	journal = {Drones},
	publisher = {Multidisciplinary Digital Publishing Institute},
	author = {Xing, Siyuan and Zhang, Xuhui and Tian, Jiandong and Xie, Chunlei and Chen, Zhihong and Sun, Jianwei},
	month = dec,
	year = {2024},
	pages = {762},
}

@inproceedings{x_morf_drone,
	title = {X-{Morf}: {A} crash-separable quadrotor that morfs its {X}-geometry in flight},
	shorttitle = {X-{Morf}},
	url = {https://ieeexplore.ieee.org/document/8101670},
	doi = {10.1109/RED-UAS.2017.8101670},
	urldate = {2026-08-13},
	booktitle = {2017 {Workshop} on {Research}, {Education} and {Development} of {Unmanned} {Aerial} {Systems} ({RED}-{UAS})},
	author = {Desbiez, A. and Expert, F. and Boyron, M. and Diperi, J. and Viollet, S. and Ruffier, F.},
	month = oct,
	year = {2017},
	pages = {222--227},
}

@article{morphology_adaptation_flight_performance,
	title = {Comparative {Validation} {Study} on {Bioinspired} {Morphology}-{Adaptation} {Flight} {Performance} of a {Morphing} {Quad}-{Rotor}},
	volume = {6},
	issn = {2377-3766},
	url = {https://ieeexplore.ieee.org/document/9398538},
	doi = {10.1109/LRA.2021.3071673},
	number = {3},
	urldate = {2026-08-13},
	journal = {IEEE Robotics and Automation Letters},
	author = {Zhao, Na and Yang, Weixin and Peng, Cong and Wang, Gang and Shen, Yantao},
	month = jul,
	year = {2021},
	pages = {5145--5152},
}

@inproceedings{sarrus_linkage_single_dof_drone,
	title = {A {Novel} {Morphing} {Quadrotor} {UAV} with {Sarrus}-{Linkage}-{Based} {Reconfigurable} {Frame}},
	url = {https://ieeexplore.ieee.org/document/10619988},
	doi = {10.1109/ReMAR61031.2024.10619988},
	urldate = {2026-08-13},
	booktitle = {2024 6th {International} {Conference} on {Reconfigurable} {Mechanisms} and {Robots} ({ReMAR})},
	author = {Wang, Yan and Liu, Chen and Zhang, Ketao},
	month = jun,
	year = {2024},
	pages = {283--289},
}

@inproceedings{transformable_linear_planar_multirotor,
	title = {Whole-body aerial manipulation by transformable multirotor with two-dimensional multilinks},
	url = {https://ieeexplore.ieee.org/document/7989606},
	doi = {10.1109/ICRA.2017.7989606},
	urldate = {2026-08-13},
	booktitle = {2017 {IEEE} {International} {Conference} on {Robotics} and {Automation} ({ICRA})},
	author = {Zhao, Moju and Kawasaki, Koji and Chen, Xiangyu and Noda, Shintaro and Okada, Kei and Inaba, Masayuki},
	month = may,
	year = {2017},
	pages = {5175--5182},
}

@article{dragon_nonplanar_transforming_drone,
	title = {Design, {Modeling}, and {Control} of an {Aerial} {Robot} {DRAGON}: {A} {Dual}-{Rotor}-{Embedded} {Multilink} {Robot} {With} the {Ability} of {Multi}-{Degree}-of-{Freedom} {Aerial} {Transformation}},
	volume = {3},
	issn = {2377-3766},
	shorttitle = {Design, {Modeling}, and {Control} of an {Aerial} {Robot} {DRAGON}},
	url = {https://ieeexplore.ieee.org/document/8258850},
	doi = {10.1109/LRA.2018.2793344},
	number = {2},
	urldate = {2026-08-13},
	journal = {IEEE Robotics and Automation Letters},
	author = {Zhao, Moju and Anzai, Tomoki and Shi, Fan and Chen, Xiangyu and Okada, Kei and Inaba, Masayuki},
	month = apr,
	year = {2018},
	pages = {1176--1183},
}

@article{ring_rotor_drone,
	title = {Ring-{Rotor}: {A} {Novel} {Retractable} {Ring}-{Shaped} {Quadrotor} {With} {Aerial} {Grasping} and {Transportation} {Capability}},
	volume = {8},
	issn = {2377-3766},
	shorttitle = {Ring-{Rotor}},
	url = {https://ieeexplore.ieee.org/document/10044964},
	doi = {10.1109/LRA.2023.3245499},
	number = {4},
	urldate = {2026-08-13},
	journal = {IEEE Robotics and Automation Letters},
	author = {Wu, Yuze and Yang, Fan and Wang, Ze and Wang, Kaiwei and Cao, Yanjun and Xu, Chao and Gao, Fei},
	month = apr,
	year = {2023},
	pages = {2126--2133},
}

@article{eagle_claw_drone,
	title = {Biomimetic {Morphing} {Quadrotor} {Inspired} by {Eagle} {Claw} for {Dynamic} {Grasping}},
	volume = {40},
	issn = {1941-0468},
	url = {https://ieeexplore.ieee.org/document/10495172},
	doi = {10.1109/TRO.2024.3386616},
	urldate = {2026-08-13},
	journal = {IEEE Transactions on Robotics},
	author = {Xu, Mengxin and De, Qixin and Yu, Dafang and Hu, An and Liu, Zhe and Wang, Hesheng},
	year = {2024},
	pages = {2513--2528},
}

@article{overactuated_quadrotor,
	title = {A {Novel} {Overactuated} {Quadrotor} {Unmanned} {Aerial} {Vehicle}: {Modeling}, {Control}, and {Experimental} {Validation}},
	volume = {23},
	issn = {1558-0865},
	shorttitle = {A {Novel} {Overactuated} {Quadrotor} {Unmanned} {Aerial} {Vehicle}},
	url = {https://ieeexplore.ieee.org/document/6868215},
	doi = {10.1109/TCST.2014.2330999},
	number = {2},
	urldate = {2026-08-14},
	journal = {IEEE Transactions on Control Systems Technology},
	author = {Ryll, Markus and Bülthoff, Heinrich H. and Giordano, Paolo Robuffo},
	month = mar,
	year = {2015},
	pages = {540--556},
}

@article{li2019particle,
	title = {Particle robotics based on statistical mechanics of loosely coupled components},
	volume = {567},
	number = {7748},
	issn = {0028-0836, 1476-4687},
	url = {https://www.nature.com/articles/s41586-019-1022-9},
	doi = {10.1038/s41586-019-1022-9},
	journal = {Nature},
	author = {Li, Shuguang and Batra, Richa and Brown, David and Chang, Hyun-Dong and Ranganathan, Nikhil and Hoberman, Chuck and Rus, Daniela and Lipson, Hod},
	month = mar,
	year = {2019},
	pages = {361--365},
}

@article{belke2017mori,
	title = {Mori: A Modular Origami Robot for Physical Locomotion and Reconfiguration},
	volume = {22},
	number = {5},
	issn = {1083-4435, 1941-014X},
	url = {https://doi.org/10.1109/TMECH.2017.2697310},
	doi = {10.1109/TMECH.2017.2697310},
	journal = {IEEE/ASME Transactions on Mechatronics},
	author = {Belke, Christoph H. and Paik, Jamie},
	month = oct,
	year = {2017},
	pages = {2153--2164},
}

@article{zhao2018affine,
	title = {Affine Formation Maneuver Control of Multiagent Systems},
	volume = {63},
	number = {12},
	issn = {0018-9286, 1558-2523},
	url = {https://doi.org/10.1109/TAC.2018.2798805},
	doi = {10.1109/TAC.2018.2798805},
	journal = {IEEE Transactions on Automatic Control},
	author = {Zhao, Shiyu},
	month = dec,
	year = {2018},
	pages = {4140--4155},
}

@article{liu2019distributed,
	title = {Distributed Reconfiguration Planning for Modular Robots With Reconfigurable Topology},
	volume = {4},
	number = {4},
	issn = {2377-3766},
	url = {https://doi.org/10.1109/LRA.2019.2930432},
	doi = {10.1109/LRA.2019.2930432},
	journal = {IEEE Robotics and Automation Letters},
	author = {Liu, Chao and Lin, Ting-Kuei and Yim, Mark},
	month = oct,
	year = {2019},
	pages = {4231--4238},
}

@article{wilcox2007athlete,
	title = {{ATHLETE}: A Cargo Handling and Manipulation Robot for the Moon},
	volume = {24},
	number = {5},
	issn = {1556-4959, 1556-4967},
	url = {https://doi.org/10.1002/rob.20193},
	doi = {10.1002/rob.20193},
	journal = {Journal of Field Robotics},
	author = {Wilcox, Brian H. and Litwin, Todd and Biesiadecki, Jeffrey and Matthews, J. and Heverly, Matthew and Morrison, Jack and Townsend, Julie and Ahmad, N. and Sirota, A. and Cooper, Brian},
	month = may,
	year = {2007},
	pages = {421--434},
}

\end{document}